\documentclass[10pt,sigconf,letterpaper,nonacm]{acmart}

\AtBeginDocument{%
  }

\setcopyright{acmlicensed}
\copyrightyear{2018}
\acmYear{2018}
\acmDOI{XXXXXXX.XXXXXXX}

\usepackage{hyperref}
\usepackage{graphicx}
\usepackage{amsmath}
\usepackage{amsfonts}
\usepackage{booktabs}
\usepackage{xcolor}
\usepackage{subcaption}
\usepackage{caption}  
\usepackage{tcolorbox}
\usepackage{mdframed}
\usepackage{multirow}
\usepackage{float}
\usepackage[table]{xcolor} 
\definecolor{darkgreen}{RGB}{0,150,0}
\usepackage{enumitem}
\usepackage{pifont}

\begin{document}
\settopmatter{printfolios=true}

\title{Quantifying the Privacy Implications of High-Fidelity Synthetic Network Traffic}


\author{Van Tran}
\orcid{0000-0002-0835-8598}
\affiliation{
    \institution{University of Chicago}
    \city{Chicago}
    \country{United States}
}
\author{Shinan Liu}
\orcid{0000-0002-6170-2167}
\affiliation{
    \institution{University of Hong Kong}
    \city{Hong Kong}
    \country{China}
}
\author{Tian Li}
\orcid{0000-0002-3735-2240}
\affiliation{%
  \institution{University of Chicago}
    \city{Chicago}
    \country{United States}
}
\author{Nick Feamster}
\orcid{0000-0001-9315-5201}
\affiliation{%
  \institution{University of Chicago}
    \city{Chicago}
    \country{United States}
}

\begin{abstract}

To address the scarcity and privacy concerns of network traffic data, various generative models have been developed to produce synthetic traffic. However, synthetic traffic is not inherently privacy-preserving, and the extent to which it leaks sensitive information, and how to measure such leakage, remain largely unexplored. This challenge is further compounded by the diversity of model architectures, which shape how traffic is represented and synthesized. We introduce a comprehensive set of privacy metrics for synthetic network traffic, combining standard approaches like membership inference attacks (MIA) and data extraction attacks with network-specific identifiers and attributes. Using these metrics, we systematically evaluate the vulnerability of different representative generative models and examine the factors that influence attack success.
Our results reveal substantial variability in privacy risks across models and datasets. MIA success ranges from 0\% to 88\%, and up to 100\% of network identifiers can be recovered from generated traffic, highlighting serious privacy vulnerabilities. We further identify key factors that significantly affect attack outcomes, including training data diversity and how well the generative model fits the training data.
These findings provide actionable guidance for designing and deploying generative models that minimize privacy leakage, establishing a foundation for safer synthetic network traffic generation.
\end{abstract}
\maketitle



%

\section{Introduction} \label{sec:intro}

Network traffic generators have gained significant traction in both academia and industry due to their versatility \cite{coull2007playing,NetSimTM,yin2022practical,jiang2024netdiffusion}. They can augment training data for underrepresented scenarios (such as network attacks), enabling more robust anomaly-detection models \cite{10.1145/3548785.3548793, sarwar2024mycroft}. They also provide a practical way to generate diverse, realistic traffic for analysis, algorithm testing, and performance benchmarking—allowing users to work with high-fidelity data without needing access to raw traffic traces \cite{coull2007playing,10.1145/3548785.3548793, van2023membership}.

However, generative modeling does not automatically come with privacy \cite{hilprecht2019monte, hayes2017logan, zhou2021property}. While prior work has focused largely on fidelity \cite{chu2025netssm, jiang2024netdiffusion, yin2022practical}, i.e., how realistic the generated traffic appears, much less attention has been paid to the privacy implications of synthetic traffic generation. In practice, many generative models either overlook privacy risks entirely \cite{chu2025netssm, cui2025trafficllm} or offer only minimal protection against them. Compounding this issue, systematic evaluation of privacy risks in synthetic network traffic remains scarce. A few existing efforts primarily rely on existing membership inference attacks (MIA) \cite{yin2022practical, sun2024netdpsyn, jin2025assessing} as the sole measure of privacy leakage, even though network traffic contains highly sensitive information beyond data membership and other network-specific privacy concerns \cite{10533453,maali2025evaluating,yao2025trafficdiary,coull2007playing}.

To bridge the gap between real-world privacy risks in network traffic and the limited metrics currently used to evaluate them, we introduce a comprehensive suite of privacy metrics spanning three complementary attack types: (1) membership inference attacks, (2) data extraction attacks, and (3) network-specific privacy attacks.
MIA and data extraction are standard privacy attacks against generative models~\cite{hu2023membership,duan2024membership,hayase2024data}. MIA evaluates whether an adversary can determine if a particular traffic sample was included in the model's training data, while data extraction attacks assess whether the model memorizes and reproduces training data. However, both methods lack interpretability in the networking context, making it difficult for dataset owners to understand what information is being exposed, whether the leaked information is sensitive, and the degree of leakage.

Leveraging insights from real-world privacy attacks against network traffic including fingerprinting attacks \cite{chen2014fingerprinting, bai2022passive}, attribute inference attacks \cite{yao2025trafficdiary, demographics} and topology reconstruction \cite{yao2003topology, hardness}, we introduce network-specific privacy metrics that capture key privacy risks in network traffic comprising (i) network identifier memorization (e.g., IP addresses, MAC addresses), which directly exposes individual users and devices; (ii) sensitive properties leakage, which reveals private user behaviors, visited websites, and application usage patterns; and (iii) topology extraction, which exposes structural relationships between network entities and can compromise organizational security. Together, these metrics provide actionable insights into the real-world privacy implications of synthetic traffic generation, enabling practitioners to make informed deployment decisions.

We evaluate our metrics across four state-of-the-art generative models, each representing a distinct generation paradigm (GAN, diffusion, transformer, SSM), and five datasets collected under diverse network conditions and tasks. Here is the summary of our research questions:

\begin{itemize}[leftmargin=*]
    \item \textbf{RQ1: How effective can each privacy attack be under the most challenging conditions for an attacker? (\textsection \ref{sec:RQ1})} We evaluate MIA, data extraction, and network-specific privacy attacks across network generative models under \textit{the most challenging conditions for an attacker: attackers only have black-box access 
    and each generative model only undergoes minimal training} (i.e., less likely to overfit to or memorize training data). We find that even under these constrained conditions, MIA attacks, particularly against language-based models (TrafficLLM and NetSSM), can achieve True Positive Rate (TPR) exceeding 0.7. Additionally, up to 35\% of network identifiers in training data can be correctly recovered.

    \item \textbf{RQ2: Which factors impact the effectiveness of privacy attacks? (\textsection \ref{sec:RQ2})} We evaluate the factors that most significantly influence each privacy attack type. These include attacker capabilities (black-box vs. white-box access), model overfitting, and data properties (training data size and diversity). Our findings reveal that for membership inference attacks, white-box access and doubling the training data size each increase the true positive rate by 0.5. For data extraction and network-specific attacks, models with greater overfitting increase the number of memorized identifiers by 0.50.

    \item \textbf{RQ3: What is the privacy-utility tradeoff when applying privacy mitigation techniques?} We explore whether mitigation strategies such as identifier anonymization and noise addition can reduce privacy risks, and examine the privacy-utility tradeoffs of these approaches. Our findings show that no single mitigation strategy can address all privacy risks. Anonymization-based methods often come at a high cost to model fidelity and downstream task utilities compared to noise-based methods.
    
    
    
\end{itemize}

The remainder of this paper is organized as follows. Section \ref{sec:relatedwork} covers related work. Section \ref{sec:method} describes privacy attack and mitigation methodology. Section \ref{sec:setup} presents experimental setup. Section \ref{sec:RQ1}, \ref{sec:RQ2}, and \ref{sec:RQ3} each covers results for a research question. Section \ref{sec:discussion} discusses implications of our findings and we conclude our work in Section~\ref{sec:conclusion}.

\section{Related Work}\label{sec:relatedwork}

This section presents related work on network traffic generation methods, privacy risks in generative models and network traffic, and privacy mitigation techniques.
\subsection{Network Traffic Generation} \label{sec:related:generation}

Synthetic trace generation has long been an important topic in network research, and consequently, a substantial body of work has been dedicated to advancing this area.

\begin{table*}[t]
    \footnotesize
    \centering
    \resizebox{0.97\textwidth}{!}{
    \begin{tabular}{llllp{9cm}}
        \toprule
        \textbf{Model} & \textbf{Base Model} & \textbf{Granularity}& \textbf{Unit} & \textbf{Generated Data} \\
        \midrule
        NetDiffusion & Diffusion&Packet & Bit & Network layer (except IP addresses), transport layer for the first 1024 packets. \\
        NetSSM & State-space model&Packet & Byte & Ethernet layer, network layer and transport layer for the entire PCAP file. \\
        TrafficLLM & Transformer &Packet& Hex digits & Ethernet layer, network layer, and transport layer for the first packet. \\
        NetShare & GAN &Flow &Tabular values & 5-tuples and header fields including packet length, type of service, identification, control flag, data offset, time to live, and inter-arrival time. \\
        \bottomrule
    \end{tabular}}
    \caption{Overview of network traffic generative models and their output representations.}
    \label{tab:models-overview}
\end{table*}

\textbf{Non-generative models:} Early network traffic generation relied on simulation tools \cite{zeng1998glomosim, henderson2008network, OMNeT, lacage2006yet, riley2003large, buss1996discrete} or statistical methods \cite{redvzovic2017ip, xu2021stan}. However, simulation tools fail to capture the complexity of real-world traffic, while statistical methods generate only limited network features and lack the fine-grained granularity required for many network analysis tasks.

\textbf{GAN-based models:} GANs were among the first deep learning models applied to network traffic generation. Notable examples include DoppelGANger \cite{lin2020using}, PacketCGAN \cite{wang2020packetcgan}, and NetShare \cite{yin2022practical}. NetShare, the state-of-the-art GAN-based model, generates flow-level statistics and specific packet header features such as IP addresses, ports, TTL, packet sizes, and inter-arrival times.

\textbf{Diffusion-based models:} Diffusion-based models \cite{zhang2024netdiff, sivaroopan2024netdiffus, jiang2024netdiffusion, GDPlan} represent network traffic as images and fine-tune diffusion models to generate synthetic traffic. NetDiffusion, a representative model of this approach, uses a text-to-image diffusion model to generate packet-level traffic headers.

\textbf{Transformer-based models:} Transformer-based models \cite{meng2023netgpt, qu2024trafficgpt, cui2025trafficllm} tokenize network traffic as hexadecimal values or bytes, then train or fine-tune transformer models to learn traffic representations. TrafficLLM, a representative model of this approach, uses a two-stage fine-tuning pipeline that enhances the model's instruction-following capabilities and traffic representation learning to generate packet-level network traffic headers.

 \textbf{State-space models:} State-space models (SSMs) \cite{chu2025netssm} model dynamic systems through hidden states that evolve over time. NetSSM tokenizes packet headers at the byte level with the addition of control tokens to represent traffic type and packet boundaries. These tokens are used to pre-train an SSM model, which then generates packet-level network traffic headers.

In this study, we examine privacy leakage in four state-of-the-art models: NetShare, NetDiffusion, TrafficLLM, and NetSSM. Each model represents a different generative paradigm (GANs, diffusion, transformers, and state-space models), enabling comprehensive assessment of privacy risks across diverse architectural approaches. Table \ref{tab:models-overview} presents a summary of key differences between network traffic generative models. More details about each method can be found in Appendix \ref{app:models}.

\subsection{Privacy Risks of Generative Models} \label{sec:related:privacy}

Generative models are susceptible to various types of privacy attacks listed below:

\textbf{Membership inference attacks:} Under MIA, an adversary can leverage generative models' tendency to overfit to their training data to differentiate between training versus non-training samples based on how the model behaves on them. In a typical MIA setting, attackers first extract observable signals and then apply a threshold or classifier to infer membership. In a white-box MIA where an attacker has access to the model's parameters and architecture, attackers can use a wider variety of signals such as model loss \cite{jagannatha2021membership, 8429311} and gradient on input samples \cite{pang2023black, hilprecht2019monte}. On the other hand, in a black-box attack where an attacker can only access the model's outputs, the types of signals they can use (such as  perplexity , confidence score) is more limited \cite{carlini2021extracting, 299573, shokri2017membership}. To overcome this limitation, attackers can train shadow models on data from a similar distribution to mimic the target model's behavior \cite{shokri2017membership, salem2018ml}, enabling them to learn membership inference patterns without direct access to the model's internals. MIA can be successful on a variety of models, including diffusion models \cite{pang2023black, duan2023diffusion, hu2023membership, matsumoto2023membership}, transformer-based models \cite{song2025mias, mozaffari2024semantic, ran2025lora, mattern2023membership, ren2025self} and GAN-based models \cite{hilprecht2019monte,shokri2017membership}.

\textbf{Data extraction attacks:} Extraction attacks concern with whether an attacker can reconstruct training data, be it text sequences \cite{carlini2021extracting, nasr2023scalable} or images \cite{carlini2023extracting}, from the model. Various studies have demonstrated that these attacks are effective against diffusion models \cite{liu2024unstoppable, carlini2023extracting}, transformer-based models \cite{yan2024protecting, carlini2021extracting, nasr2023scalable}, and GANs \cite{zhang2020secret, hu2021model, Tinsley_2021_WACV}. For example, leveraging the intuition that memorized samples appear with disproportionately high frequency and confidence scores, attackers prompt the model to generate a large number of outputs, rank them based on these metrics, and then examine them for memorization. Attackers have successfully extracted text sequences containing sensitive information such as names, phone numbers, and email addresses \cite{carlini2021extracting, nasr2023scalable}, as well as images of identifiable faces and company logos \cite{carlini2023extracting}.

\textbf{Other privacy attacks: }Generative models are also vulnerable to additional privacy risks, including attribute inference and linkability attacks. Attribute inference attacks \cite{zhou2021property,ganju2018property, fredrikson2015model, hayase2024data, wang2024property} aim to uncover a specific sensitive attribute of the model’s training data such as the data composition of the training data, or whether samples of a certain properties are included in the training samples. In contrast, a linkability attack \cite{powar2023sok, giomi2022unified} does not attempt to reconstruct data or infer particular attributes. Instead, it focuses on whether an adversary can determine if two records—whether generated samples or external data—originate from the same underlying entity.

\subsection{Privacy Risks of Network Traffic} \label{sec:related:ours}

Network traffic is subject to various privacy attacks that can compromise user confidentiality. Fingerprinting attacks are among the most widely studied and prevalent. By analyzing packet sizes, timing patterns, TTL values, and window sizes, adversaries can identify specific devices \cite{10533453,maali2025evaluating}, operating systems \cite{chen2014fingerprinting, bai2022passive}, and the websites users visit \cite{oh2017p, mavroudis2023adaptive}. Beyond fingerprinting, attribute inference attacks \cite{yao2025trafficdiary, demographics} enable attackers to extract information from training datasets, including demographic characteristics of the Internet users such as age group and career stage. Additionally, considerable research \cite{coull2007playing, yao2003topology, hardness} has explored methods to deanonymize networks and reconstruct network topologies from anonymized traffic data.

While privacy attacks have been extensively studied for real network traffic, privacy risks in synthetic network traffic remain largely unexplored. Jin et al. \cite{jin2025assessing} provide early evidence of such vulnerabilities, demonstrating that MIA at the traffic source level are feasible. However, their work focuses solely on MIA and does not investigate other privacy threats, particularly network-specific attacks such as fingerprinting or attribute inference. Whether MIA effectiveness correlates with other privacy attacks remains unexplored. This represents a significant gap in our understanding of privacy risks in network traffic generative models, especially regarding vulnerabilities unique to network data.

\subsection{Privacy Mitigation for Network Traffic} 

\begin{figure*}[t]
    \centering

  \includegraphics[width=0.7\linewidth]{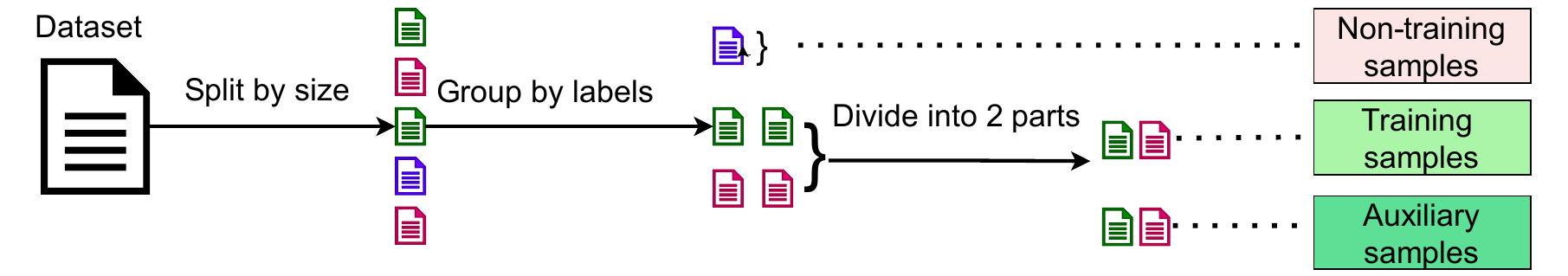}
    \caption{Pipeline of splitting each dataset into non-training, training and auxiliary data.}\label{fig:data-pipeline}
\end{figure*}

\begin{figure*}[t]
    \centering

  \includegraphics[width=0.75\linewidth]{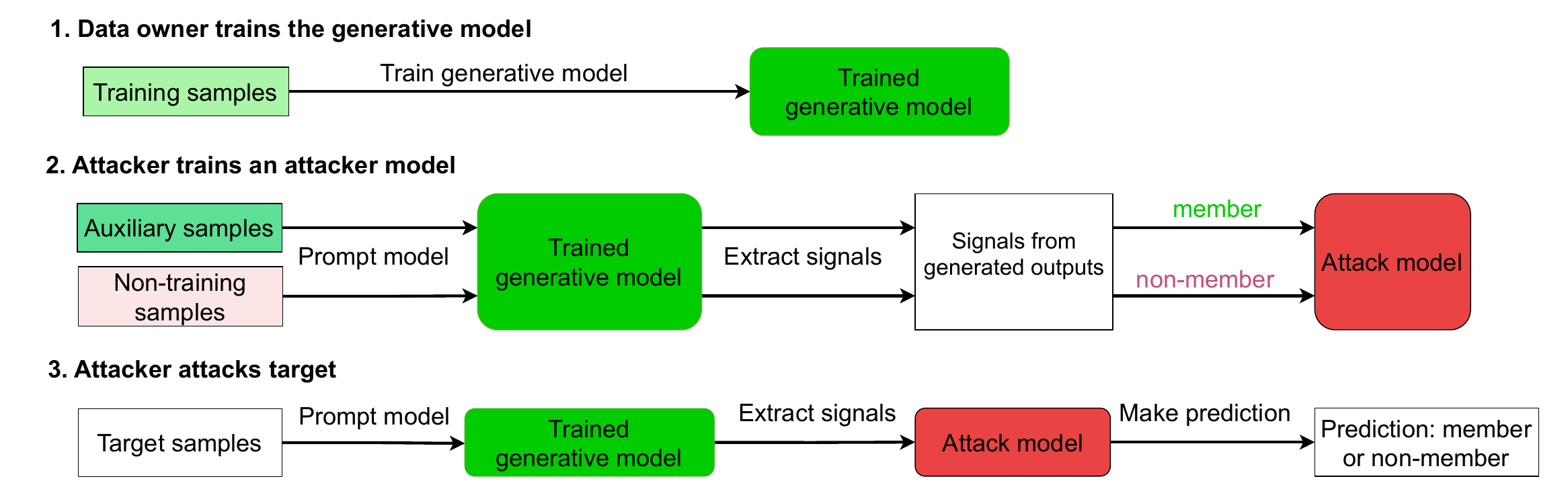}
    \caption{Pipeline of Membership Inference Attacks (MIA). (1) The data owner trains the generative model and makes it available as a service to everyone. (2) An attacker queries the generative model to extract signals and trains a binary attacker model to predict membership. (3) To predict membership of a target sample, an attacker first queries the generative model to extract signals, then predicts membership using the trained attacker model.}
    \label{fig:MIA-pipeline}
\end{figure*}

Network traffic encodes rich information about identifiers and attributes, which has motivated the development of various privacy mitigation techniques. Anonymization methods \cite{ammar2002prefix, 10.1145/1519144.1519147, slagell2006flaim}—such as prefix-preserving and reversible pseudonymization—mask network identifiers like IP and MAC addresses to prevent adversaries from recovering original values. Beyond anonymization, researchers have introduced complementary techniques to protect sensitive fields, including IP addresses, port numbers, flow sizes, and packet inter-arrival times. These approaches include reducing the granularity of packet or flow metadata \cite{aleroud2021anonymization}, perturbing feature distributions through noise addition \cite{osti_1836674, mcsherry2010differentially}, and modifying packet traces via traffic shaping \cite{chen2018taranet, cai2014cs, wang2017walkie}. 

Despite the abundance of privacy mitigation techniques for real network traffic, synthetic network traffic remains largely unprotected. Existing generative models for network traffic either lack privacy safeguards entirely \cite{cui2025trafficllm, chu2025netssm} or employ only superficial protections, such as removing IP addresses \cite{jiang2024netdiffusion} or adding differential privacy noise to defend against membership inference attacks \cite{sun2024netdpsyn, yin2022practical}. Critically, none of these approaches address network-specific privacy vulnerabilities. This gap—the lack of both comprehensive privacy protections and robust evaluation metrics for real-world privacy risks—poses significant security challenges for synthetic network traffic generation.

\section{Methods to Evaluate Privacy \& Utilities}\label{sec:method}

In Section~\ref{sec:privacy-measure-method}, we describe our methods for evaluating privacy risks through membership inference attacks, data extraction, and network-specific privacy attacks, followed by reasons why we choose to conduct these privacy attacks. Next, we explain  techniques we explore to mitigate privacy risks and methods for assessing the utility of generated traffic in downstream tasks to measure the privacy-utility trade off of these mitigation methods (Section~\ref{sec:mitigation}).

Note that not every privacy attack can be conducted against all generative models due to variations in their architectures and outputs. For example, NetDiffusion removes all network identifiers from training data, precluding identifier-based and topology-based attacks. Similarly, TrafficLLM generates only the first packet of each flow rather than complete network flows, making attribute-based network-specific attacks infeasible. Accordingly, we conduct privacy attacks against each generative model only when applicable. Table \ref{tab:privacy-eval} provides details on which privacy attacks are conducted against each model.

\subsection{Privacy Attack Methods}\label{sec:privacy-measure-method}

\subsubsection{Membership Inference Attack (MIA)}

We conduct MIA at the PCAP level, which is commonly the generation unit for network traffic models.
To launch MIA, we need to curate a set of signals (features) to train binary classifiers predicting whether a given sample is in the training set or not. Under black-box MIA, we assume the adversary only has access to the generated outputs and no access to the model's architecture or parameters. This reflects the typical setting for many deployed systems that are available only as a service, and no information is given to the attacker about the model's architecture, parameters, or training methods. Furthermore, we also assume that the attacker \textit{does not train any shadow model for MIA}. Instead, they only use signals generated along with synthetic traffic, or those that can be easily computed from the synthetic traffic. Specifically, we use pixel-wise $L_2$ reconstruction error for NetDiffusion and perplexity on the first packet's tokens for NetSSM and TrafficLLM. 
Under white‑box MIA, where an attacker can access model internals, we use the model’s loss on input samples and the average gradient norm of its parameters as signals.

After extracting signals from the non-training (labeled as non-members) and auxiliary samples (labeled as members), we train an attack model to predict the membership status of target samples. We explore several classifiers as attack models, including neural networks, support vector machines (SVMs), and XGBoost, and select the best-performing models and hyperparameters based on AUC scores. Following standard practice for MIA attacks \cite{zhu2025fedmia, kowalczuk2025privacy, carlini2022membership}, we report the true positive rate at a stringent false positive rate of 0.01 (TPR@FPR$\le$0.01) and AUC-ROC scores on the target set, which consists of both training and non-training samples. Figure \ref{fig:data-pipeline} illustrates how the non-training, training, and auxiliary samples are obtained, and Figure \ref{fig:MIA-pipeline} shows the entire MIA pipeline using these subsets of data.

\subsubsection{Data Extraction Attack}
Inspired by Nasr et al.~\cite{nasr2023scalable}, who define a successful extraction as one where the model outputs text containing a substring of at least 50 tokens that appears verbatim in the training set, we adapt this approach to the networking context as follows.
First, instead of using a 5-token block prefix from the training data as the generation prompt, we define the "prompt" as any input required by each model to generate synthetic traffic. For NetDiffusion, the prompt is the combination of an input image from the training dataset and a descriptive text prompt specifying the traffic type. For NetShare, it is the raw network traffic. For NetSSM and TrafficLLM, the prompt is simply the label of the traffic type present in the training data.

Second, we compare each generated sequence with the training data to identify verbatim matches. We define a generated sequence as extractable if it contains a \textit{consecutive sequence of at least 10 tokens} that appears verbatim in the training data. Since different generative models use different representation units—bits for NetDiffusion, bytes for NetSSM, and hexadecimal digits for TrafficLLM—we define a "token" as the native representation unit of each respective model.

\subsubsection{Network-Specific Attack}\label{sec:network-specific-attack}

Network traffic contains diverse information that varies in sensitivity. Network identifiers such as MAC addresses and IP addresses are highly sensitive, as they can reveal the location, device type, and even the identity of users. Other features, such as Time To Live (TTL) and Type of Service (ToS), while less sensitive than network identifiers, can be used to generate fingerprints that infer the operating system or application type. Furthermore, overall traffic patterns can be collectively analyzed to reconstruct network topology, exposing critical network infrastructure. To address these diverse privacy concerns, we categorize this sensitive information into three types: network identifiers, network properties, and network topology, and we evaluate each separately.

\textbf{Network Identifiers: } We propose two complementary metrics for measuring identifier leakage. \textit{Output coverage} is the percentage of training data identifiers that appear in generated outputs. \textit{Output confidence} is the probability that a generated identifier exists in the training data rather than being hallucinated. These metrics capture different memorization risks: high confidence with low coverage indicates repeated generation of a small identifier subset, while high coverage with low confidence indicates broad generation where only some identifiers are real.

\begin{equation}
\textit{Output Coverage} = 
\frac{|\{\, i \mid i \in \mathcal{G} \cap \mathcal{T} \,\}|}{|\{\, i \mid i \in \mathcal{T} \,\}|} \nonumber
\end{equation}

\begin{equation}
\textit{Output Confidence} = P(i \in \mathcal{T} \mid i \in \mathcal{G}) \nonumber
\end{equation}
where $\mathcal{G}$ denotes the set of unique identifiers generated by the model, and $\mathcal{T}$ denotes the set of unique identifiers present in the training data.

\textbf{Network properties:} Based on features used in fingerprinting and attribution inference attacks \cite{chen2014fingerprinting, bai2022passive, yao2025trafficdiary, demographics}, we select the following as sensitive network properties: Time To Live (TTL), IP Identification (IP ID), Type of Service (ToS), Window Size, Control Flags, TCP Data Offset, flow size, and packet size. Since ground truth for fingerprinting attacks is unavailable, we evaluate leakage by comparing the distributions of these features between training and generated data using normalized Earth Mover's Distance (EMD) \cite{710701}. A smaller distance indicates higher leakage risk.

\textbf{Network topology:} Network topology refers to the structure of how devices are connected and communicate in a network, including link-level connections (using MAC addresses) and network-level routing (using IP addresses). This information is highly sensitive \cite{1268075} because it exposes relationships among devices and network segments, allowing attackers to identify critical nodes and potential attack paths \cite{zhang2000detecting, soltani2024security}. To measure topology leakage, we employ a graph-based comparison between training and generated data. We define a communication flow as an exchange of network traffic (unidirectional or bidirectional) between a pair of addresses (IP or MAC). We select the top 100 communication flows ranked by total bytes transferred and construct non-directed graphs where nodes represent IP or MAC addresses. We evaluate topology similarity using three metrics: (1) node overlap—the proportion of addresses common to both graphs, (2) edge overlap—the proportion of shared communications, and (3) node connectivity distribution—the similarity in the number of connections per node across both topologies (measured using normalized EMD).

\subsubsection{Why do These Attacks Matter?}
We focus on membership inference, data extraction, and network-specific privacy attacks because they correspond to the main questions that network operators and data owners are concerned about when releasing synthetic traffic. First, MIA captures \emph{participation privacy}: can an adversary decide whether traffic from a given user, device, or organization was used to train the model? This has been widely adopted as a standard way to quantify privacy risk in machine learning~\cite{shokri2017membership} and, in the context of network traffic, directly relates to sensitive questions such as whether a host communicated with a particular service or appeared in a specific trace collection.

Second, data extraction attacks measure \emph{content memorization}: does the model reproduce concrete packet sequences from its training data when it is queried by an adversary? Prior work has shown that generative models can regurgitate individual training records~\cite{carlini2021extracting,nasr2023scalable}. For network traffic, such memorization can leak packet payloads, protocol headers, or flow patterns that operators expect to remain hidden, even when they only publish synthetic traces.

Third, our network-specific attacks target \emph{semantic leakage} that is specific to networking and not visible to generic token-level tests. By separately measuring leakage of network identifiers (such as IP and MAC addresses), fingerprinting-relevant fields (such as TTL, window size, or flags), and communication topology, these attacks align with concrete operational harms: device and user re-identification, inference of software stacks through fingerprinting, and extraction of internal network structure.

Together, these three attack families allow us to quantify privacy leakage at three levels that are directly relevant for synthetic network traffic: whether a participant's traffic was used for training (MIA), whether particular traces are memorized and reproduced (data extraction), and whether sensitive network semantics are exposed (network-specific attacks). We therefore argue that they form a meaningful core for evaluating privacy risks of traffic generative models.

\subsection{Privacy Mitigation Methods} \label{sec:mitigation}
Inspired by privacy mitigation techniques for real network traffic \cite{ammar2002prefix, 10.1145/1519144.1519147, slagell2006flaim, osti_1836674, mcsherry2010differentially}, we explore several privacy mitigation strategies and evaluate their privacy-utility tradeoffs on synthetic traffic. For each strategy, we apply mitigation on the traffic generators' training data.

\subsubsection{Anonymization-Based Strategies}
To protect network identifiers and network topology, we investigate three anonymization strategies with varying levels of restrictiveness.

\begin{itemize}
\item \textbf{Complete Anonymization (CA):}
In this most restrictive anonymization approach, each network identifier (IP addresses and MAC addresses) is replaced with a single constant value. This approach effectively eliminates all identifier-based privacy risks, including those that enable network topology reconstruction.

\item \textbf{Pseudonymization (PS):}
As a less restrictive alternative, each unique network identifier (for both IP addresses and MAC addresses) in the training data is consistently replaced with a distinct randomly generated synthetic identifier. This one-to-one mapping preserves network structure and communication patterns while obscuring the original identifiers.

\item \textbf{Prefix-preserving (PP):}
For this approach, the last k bits of each network identifier are set to zero (k=8 bits for IP addresses, k=12 bits for MAC addresses), while the rest remain intact, allowing the hierarchical prefix information to be preserved. For IP addresses, subnet structure is retained while host identities is obscured. For MAC addresses, the Organizationally Unique Identifier that identifies the manufacturer is preserved while device-specific bits are removed.
\end{itemize}

\subsubsection{Noise-Based Strategies} Adding noise to the training data, generated data, or to the generative model during training is an effective approach to protect privacy. Since most network traffic generative models do not currently support adding DP-noise directly to the model (e.g., via DP-SGD), we apply DP-noise to the training data, as in previous works ~\cite{sun2024netdpsyn}. Rather than perturbing every header field, we specifically target the sensitive network properties identified in Section \ref{sec:network-specific-attack}, allocating an equal privacy budget across all targeted properties and add noise simultaneously for all selected properties. We then apply clipping and rounding operations to ensure the perturbed values remain within valid ranges and conform to the appropriate data types.

\subsubsection{Privacy-Utility Tradeoffs Evaluation}

We investigate privacy-utility tradeoffs for both multi-flow and single-flow generation to understand how mitigation techniques affect different traffic generation tasks. We measure utility using two types of metrics: fidelity metrics that assess how well generated traffic preserves key distributions, and downstream task metrics that measure performance on practical applications. To assess privacy implications, we select key metrics from Section~\ref{sec:privacy-measure-method} covering all three attack types.

\textbf{Fidelity metrics.}
Following the fidelity metrics used in NetShare \cite{yin2022practical}, we measure distributional distance between training and generated datasets using Earth Mover's Distance (EMD) across five key distributions. \textit{An increase in EMD indicates a decrease in fidelity}, meaning the generated data diverges further from the training distribution, while \textit{a decrease in EMD indicates an increase in fidelity}.
The five evaluated distributions are:
\begin{itemize}
\item \textbf{SA/DA:} Relative frequency ranking of source and destination IP addresses (from most to least frequent)
\item \textbf{SP/DP:} Distribution of source and destination port numbers
\item \textbf{PR:} Relative frequency of IP protocols (e.g., TCP, UDP, ICMP)
\end{itemize}

\textbf{Downstream task metrics.} We use downstream task (classification) accuracies to measure the utility of synthetic traffic.\footnote{Since these downstream tasks require flow-level labels, which are only available for the IoT and SR datasets, we restrict the single-flow generation evaluation to these two datasets.} The downstream task varies by dataset based on the relevant applications. For the IoT dataset, the downstream task is device type classification, as device identification is critical for anomaly detection and security monitoring \cite{nguyen2019diot, meidan2017detection}. For the Service Recognition (SR) dataset, the downstream task is service type prediction (e.g., video-conferencing, streaming), which is essential for QoS provisioning and network resource allocation \cite{aziz2023content}.

To assess how mitigation techniques impact downstream task performance, we train two XGBoost classifiers: one on unmitigated synthetic traffic and another on mitigated synthetic traffic. Both classifiers are evaluated on the same real-world test set (auxiliary data from Figure \ref{fig:data-pipeline}). The difference in classification accuracy between these two models quantifies the utility cost of applying mitigation techniques.
\begin{table}[t]
    \footnotesize
    \centering
    \resizebox{0.45\textwidth}{!}{
    \begin{tabular}{l|p{1cm}|p{1.75cm}|p{1cm}p{1cm}p{1cm}} 
    \toprule

       & \textbf{MIA} & \textbf{Data} & \multicolumn{3}{c}{\textbf{Network-specific}} \\
       & & \textbf{extraction}&\textbf{Identifiers} & \textbf{Attributes} & \textbf{Topology}\\
       \midrule
       NetShare& \textcolor{red}{\ding{55}} & \textcolor{red}{\ding{55}}  & \textcolor{darkgreen}{\checkmark} & \textcolor{darkgreen}{\checkmark} &\textcolor{darkgreen}{\checkmark}\\
       NetDiffusion&\textcolor{darkgreen}{\checkmark} & \textcolor{darkgreen}{\checkmark} & \textcolor{red}{\ding{55}} & \textcolor{darkgreen}{\checkmark}& \textcolor{red}{\ding{55}} \\
       TrafficLLM& \textcolor{darkgreen}{\checkmark}& \textcolor{darkgreen}{\checkmark}& \textcolor{darkgreen}{\checkmark}& \textcolor{red}{\ding{55}} & \textcolor{red}{\ding{55}} \\
       NetSSM& \textcolor{darkgreen}{\checkmark}&\textcolor{darkgreen}{\checkmark} &\textcolor{darkgreen}{\checkmark} &\textcolor{darkgreen}{\checkmark} &\textcolor{darkgreen}{\checkmark}\\

    \bottomrule   
    \end{tabular}
    }
    \caption{Privacy attacks conducted for each generative model. \textcolor{darkgreen}{\checkmark} indicates an attack was conducted; \textcolor{red}{\ding{55}} indicates "not conducted" because of different data encodings and model design choices.}\label{tab:privacy-eval}
    
\end{table}

\textbf{Privacy metrics.} To examine how mitigation techniques impact each privacy attack type, we select one representative metric per attack category. For membership inference attacks, we use $\Delta$ TPR at FPR $\le$ 0.01. For data extraction attacks, we use $Delta$ extractable rate. For network-specific attacks, we measure privacy leakage of sensitive network properties using normalized EMD between training and generated distributions. Lower EMD indicates more effective attacks (closer distributional match enables easier property inference), while higher EMD indicates improved privacy protection. To maintain consistency with other metrics where positive changes indicate improved privacy, we report the change in network-specific attack effectiveness as the negative of the change in EMD (i.e., -$\Delta$EMD), averaged across all sensitive network properties and datasets.
\section{Experimental Setup}\label{sec:setup}

\begin{table}[t]
    \centering
    \resizebox{0.48\textwidth}{!}{
    \setlength{\tabcolsep}{3pt}
    \begin{tabular}{lrrrr}
    \toprule
        \textbf{Traffic Dataset}&  \textbf{\# Pcaps} & \textbf{\# Flows} & \textbf{\# Packets}  & \textbf{\# Labels}\\
    \midrule
        IoT (IoT)&  521  & 9,023 & 1,007,501 & 9\\
        VNAT (VNAT)&  1103  & 14,646& 2,118,603&19\\
        Service recognition (SR)& 660   & 2,363 & 1,313,372 & 11\\
        
        USTC TFC 2016 (USTC)& 501  & 183,570 &  994,174 & 13\\
CIC DoHBrw 2020 (CIC)&   480 & 10,758  &  838,140 & 8\\

        \bottomrule
         
    \end{tabular}}
    \caption{Overview of five network traffic datasets.}
    \label{tab:datasets-description}
\end{table}

This section describes the target to be generated in our experiments, the dataset preparation process, and the details of synthetic traffic generation.

\subsection{Generation Target}
In this work, we generate synthetic network traffic for \textit{multi-flow PCAPs}.

Unlike single-flow PCAPs that capture communication between two entities, multi-flow PCAPs contain multiple concurrent flows, each defined by a 5-tuple (source IP, destination IP, source port, destination port, protocol).
Multi-flow traffic generation is valuable for practical applications including network traffic analytics, monitoring, and simulating realistic environments to test network algorithms and protocols. However, it presents unique technical and privacy challenges that make it particularly important to study.

From a technical perspective, multi-flow generation is more complex than single-flow generation. While single-flow generation only needs to model the temporal dynamics within one communication session (intra-flow patterns), multi-flow generation must additionally capture the relationships and dependencies between different flows (inter-flow patterns), such as timing correlations and bandwidth sharing across concurrent connections. From a privacy perspective, multi-flow traffic exposes more sensitive information than single-flow traffic. Beyond individual communication details, multi-flow data reveals network-wide communication patterns, entity relationships, and network topology—information that is highly sensitive and valuable to adversaries. These elevated privacy risks make multi-flow traffic generation a critical area for privacy evaluation.

Given both the practical importance and heightened privacy concerns of multi-flow traffic generation, we focus primarily on this scenario throughout our evaluation.

\subsection{Dataset Preparation}\label{sec:data-curation}

We use five datasets collected from different networks and application domains. The IoT dataset~\cite{liu2023amir} contains traffic from various IoT device activities. The Service Recognition (SR)~\cite{jiang2024netdiffusion} and VNAT~\cite{vnat_dataset} datasets capture traffic from different service types, including streaming, conferencing, and social media. The USTC TFC 2016 (USTC)~\cite{ustc} and CIC DoHBrw 2020 (CIC)~\cite{cic} datasets provide network traffic from various attack scenarios.

Since packet captures (PCAPs) vary widely in length—with some exceeding 100,000 packets—we split each PCAP into smaller segments of up to 2,000 packets and randomly select at most 20 segments from each original PCAP. Additionally, since some labels contain significantly more PCAPs than others, we limit each label to a maximum of 60 PCAPs to ensure balanced representation across labels. Table \ref{tab:datasets-description} provides details about each dataset after this curation process.

For each dataset, we first group PCAPs by label. We then select two labels and set aside all PCAPs from those labels as non-training samples (to be used in MIA attack). The PCAPs associated with the remaining labels are partitioned into two equal parts: one part is used to train the generative models, and the other serves as auxiliary data (to be used in MIA attack). Refer to Figure \ref{fig:data-pipeline} for how each dataset is split.

\subsection{Synthetic Traffic Generation}

After training traffic generative models, we produce synthetic traffic as follows. More details about each model can be found in Appendix~\ref{app:models}.

For NetDiffusion, we generate 20 output images for each input image (where each image represents a multi-flow PCAP file). We select the output closest to the input based on pixel-wise L2 distance and apply the post-generation correction process to the selected image. For NetSSM and TrafficLLM, we generate 30 output sequences per traffic label prompt (e.g., "Netflix"). This matches the approximate number of PCAPs per label in the training data, ensuring balanced sample representation across labels. For NetShare, we use the default configuration without DP noise to enable fair comparison of privacy leakage across all generative models. Following NetShare's training procedure, we merge all PCAPs with the same label into a single PCAP, which is then used to synthesize traffic.

\section{How effective can each privacy attack be under the most challenging conditions for an attacker?}\label{sec:RQ1}

To answer RQ1, we start from a conservative setting where the attacker is limited to black‑box access and the generative models (except NetDiffusion) are trained minimally for only one epoch to reduce the risk of of overfitting. We then mount the membership inference, data extraction, and network‑specific attacks defined in Section 3.

\begin{figure}[t]
    \centering

    \begin{subfigure}[b]{\linewidth}
        \centering
        \includegraphics[width=0.67\linewidth]{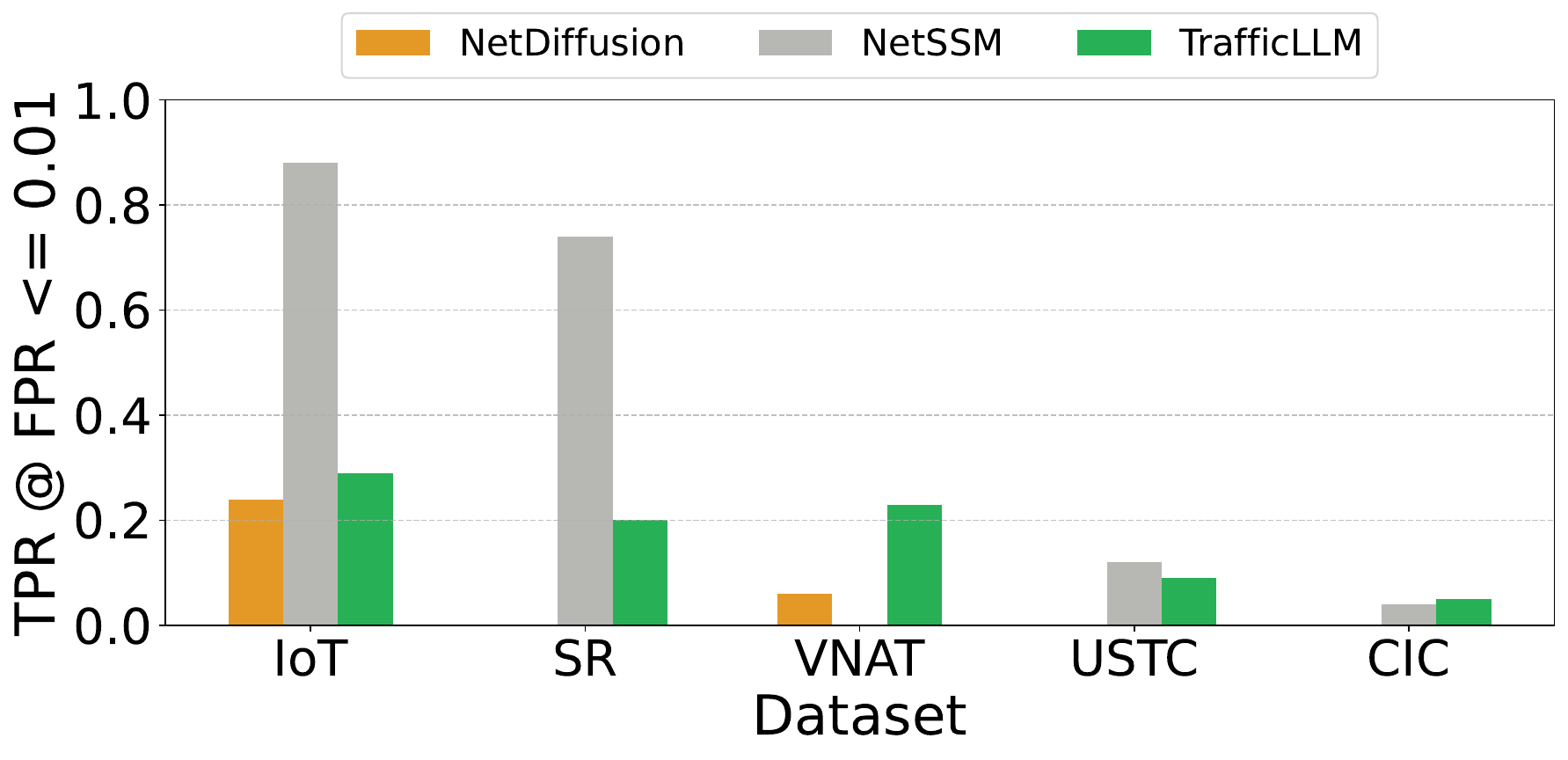}
        \caption{TPR at FPR $\le$ 0.01}
      
    \end{subfigure}
    \hfill
    \begin{subfigure}[b]{\linewidth}
        \centering
        \includegraphics[width=0.67\linewidth]{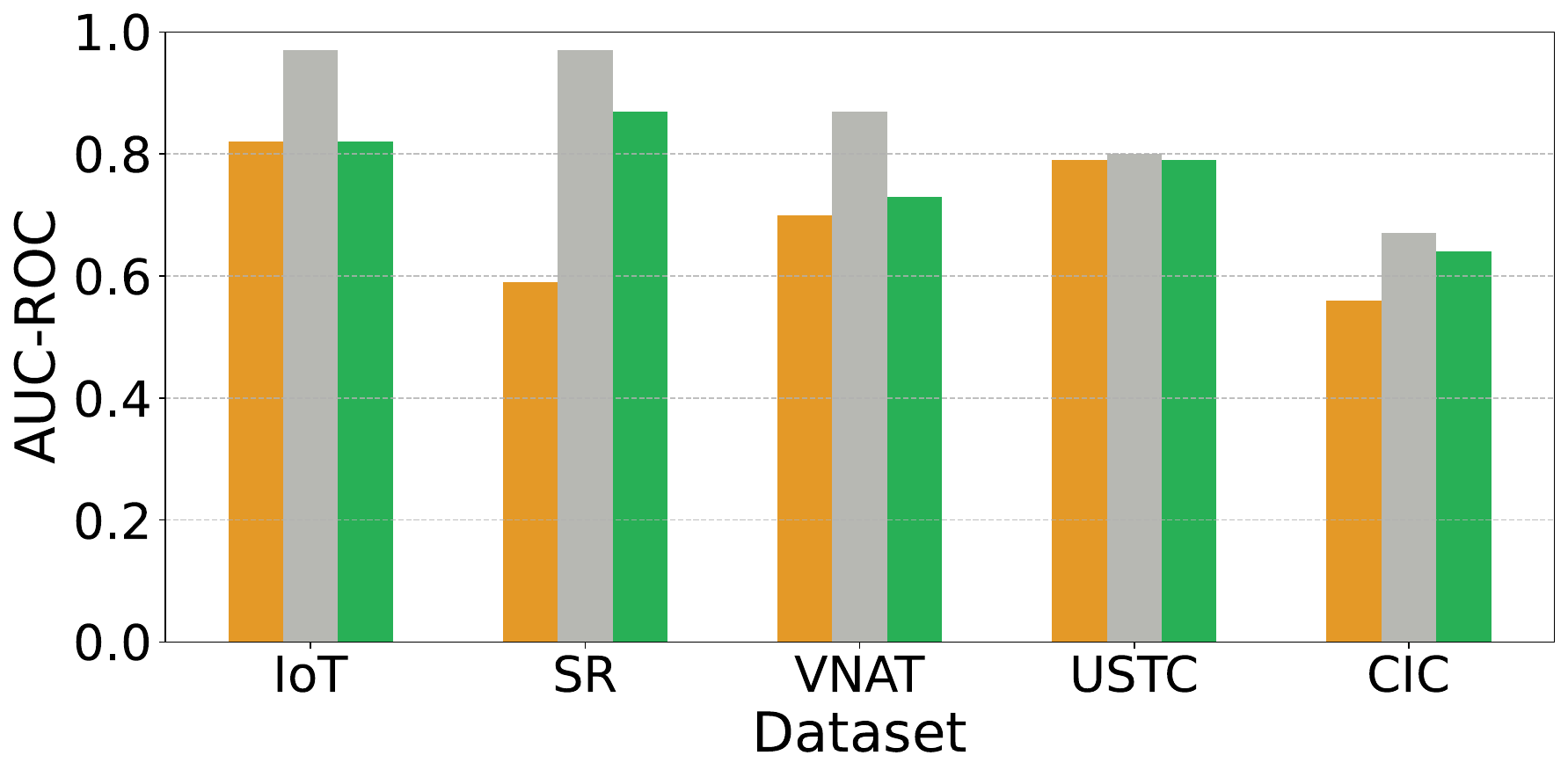}
        \caption{AUC-ROC}
       
    \end{subfigure}

    \caption{MIA performance across different generative models and datasets. NetSSM is most vulnerable, followed by TrafficLLM and NetDiffusion. MIA is highly successful on IoT and SR datasets.}
    \label{fig:mia_basecase}
\end{figure}

\subsection{MIA Success Rate}

\newcounter{findingcounter}

\newcommand{\finding}[1]{
    \refstepcounter{findingcounter}
    \noindent\rule{\columnwidth}{1.5pt}
    
    \noindent\textbf{Finding \thefindingcounter: #1}
    
    \noindent\rule{\columnwidth}{1.5pt}
}

\finding{MIA can be highly effective even under minimal attacker knowledge.}

MIA achieves an AUC $\ge$ 0.80 on four out of five datasets. Even under the strict constraint of FPR $\le$ 0.01, TPR reaches up to 0.88. MIA is highly effective against NetSSM, moderately effective against TrafficLLM, but considerably less effective against NetDiffusion (Figure~\ref{fig:mia_basecase}). 

We hypothesize that MIA success rates vary across models due to differences in data encoding and architecture. Byte-level (NetSSM) or hexadecimal (TrafficLLM) encoding likely provides more informative representations than bit-level encoding (NetDiffusion), as many network header fields such as ToS, ID, and IP addresses have lengths that are multiples of 4 or 8 bits, which are more likely to be diffused during the generation process. Furthermore, language models are better suited to capture the inherently sequential structure of network traffic than the NetDiffusion architectures, which lack native sequence modeling capabilities.

Furthermore, MIA effectiveness also varies across datasets. One possible explanation is the degree of similarity between training and non-training data: when training samples closely resemble non-training samples, the distinction between the two becomes less pronounced, potentially reducing MIA effectiveness.

\subsection{Data Extraction Attack Success Rate}

\newcommand{\findingtwo}[1]{
    \noindent\rule{\columnwidth}{1.5pt}
    
    \noindent\textbf{Finding 2: #1}
    
    \noindent\rule{\columnwidth}{1.5pt}
}

\findingtwo{Data extraction attack is more successful against autoregressive models than diffusion models. Categorical/identifier data are more likely to be memorized than continuous data.}

\begin{figure}[t]
    \centering

  \includegraphics[width=0.70\linewidth]{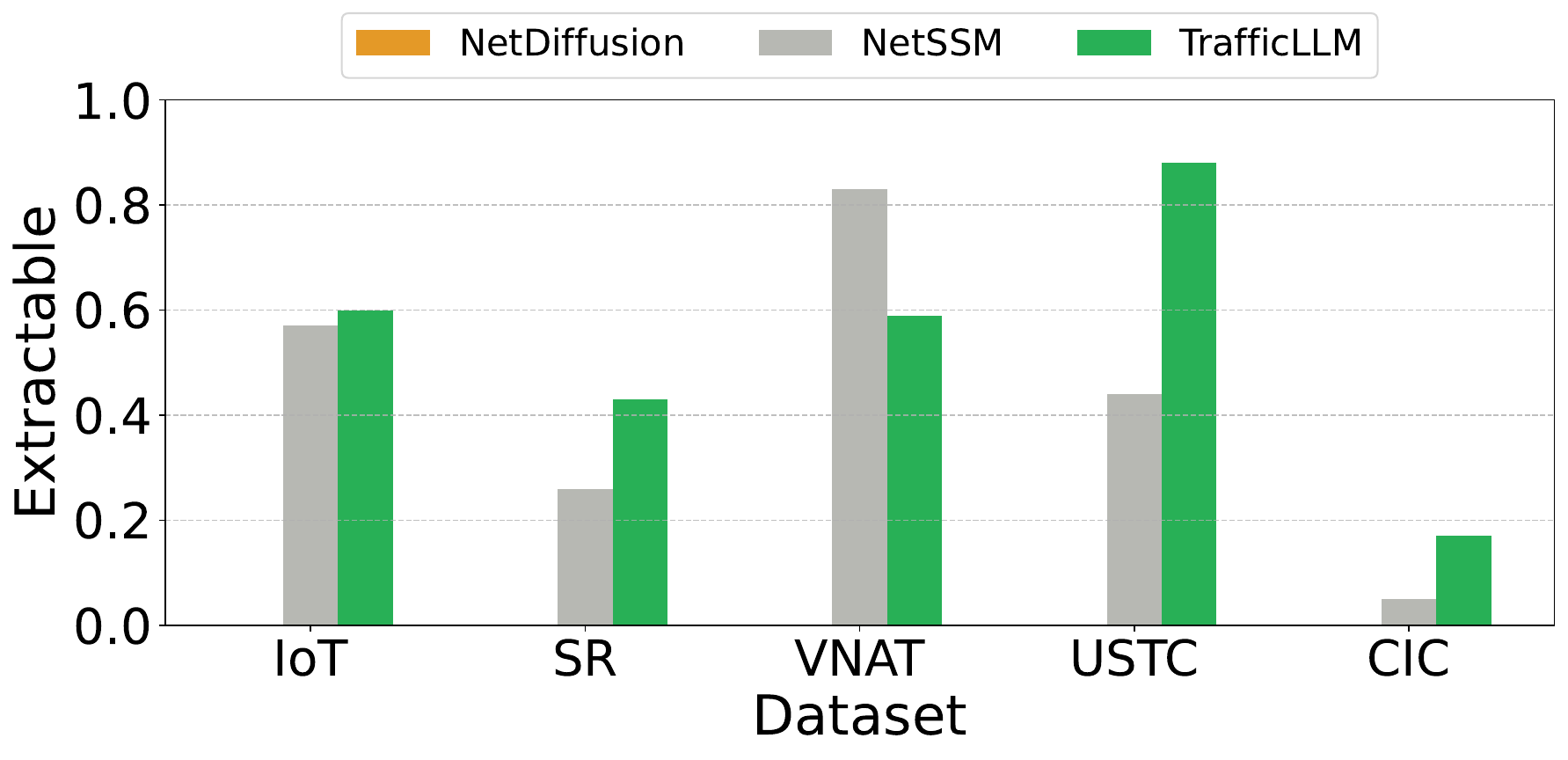}
    \caption{Data extraction attack performance across different generative models and datasets. NetDiffusion shows no vulnerability to data extraction, while extractable rates on NetSSM and TrafficLLM can exceed 0.8 on certain datasets.}\label{fig:extractable}
\end{figure}

\begin{figure}[t]
    \centering

  \includegraphics[width=0.70\linewidth]{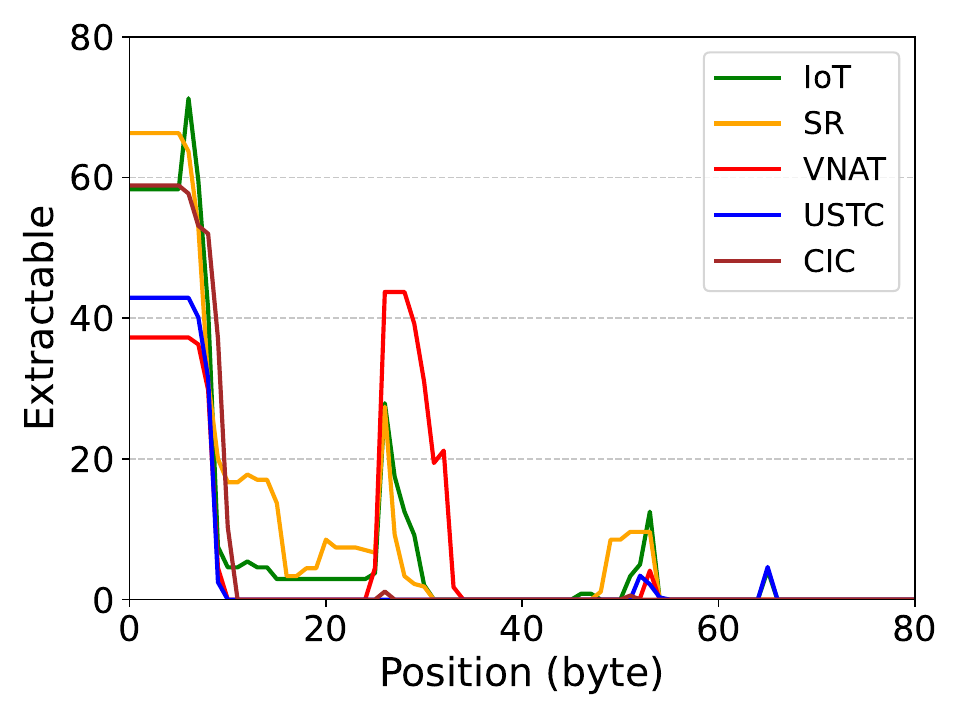}
    \caption{Extractable rate across different positions for NetSSM. Higher extractable rate is observed at positions 1-9 and 27-33.}\label{fig:extractable-position-NetSSM}
\end{figure}

Among the three packet-level generative models, NetDiffusion is not vulnerable to data extraction attacks. This aligns with the design choice in NetDiffusion: representing traffic as images at bit granularity removes explicit sequence structure, which likely makes verbatim packet‑level memorization harder to sustain. In contrast, the extractable rate is notably high ($\ge$ 0.4) across four datasets (except CIC dataset) for both NetSSM and TrafficLLM even though these models use only traffic labels (no traffic data) to prompt the model. This demonstrates significant privacy concerns: attackers can extract training samples with a high success rate even without knowledge of the training data. Refer to Figure \ref{fig:extractable} for more details about data extraction effectiveness.

Next, we examine which positions are most susceptible to extraction. We define a position as extractable if the next 10 tokens at that position appear identically in the training data at the same location (see Figure~\ref{fig:extractable-position-NetSSM} for NetSSM; Figure~\ref{fig:extractable-position-TrafficLLM} in the Appendix for TrafficLLM). The extractable rate curves show consistent patterns across datasets: extraction rates are high at positions 1--9 and 27--33, which correspond to identifier fields such as MAC and IP addresses. For instance, positions 1--9 (bytes 1--19 in NetSSM) map to the Ethernet layer (14 bytes), including source and destination MAC addresses (6 bytes each) and EtherType. These categorical fields appear frequently at fixed positions, making them easier to be memorized. In contrast, fields with continuous values are memorized less effectively due to their lower frequency and variability. Consequently, data extraction attacks succeed at some positions but fail at others, reflecting the mixed categorical and continuous nature of network headers.

\subsection{Network-specific Attack Success Rate}

\newcommand{\findingthree}[1]{
    \noindent\rule{\columnwidth}{1.5pt}
    
    \noindent\textbf{Finding 3: #1}
    
    \noindent\rule{\columnwidth}{1.5pt}
}

\findingthree{Most generative models leak network identifiers and sensitive network properties. NetSSM additionally leaks network topology.}

\begin{figure}[t]
    \centering

    \begin{subfigure}[b]{\linewidth}
        \centering
        \includegraphics[width=0.75\linewidth]{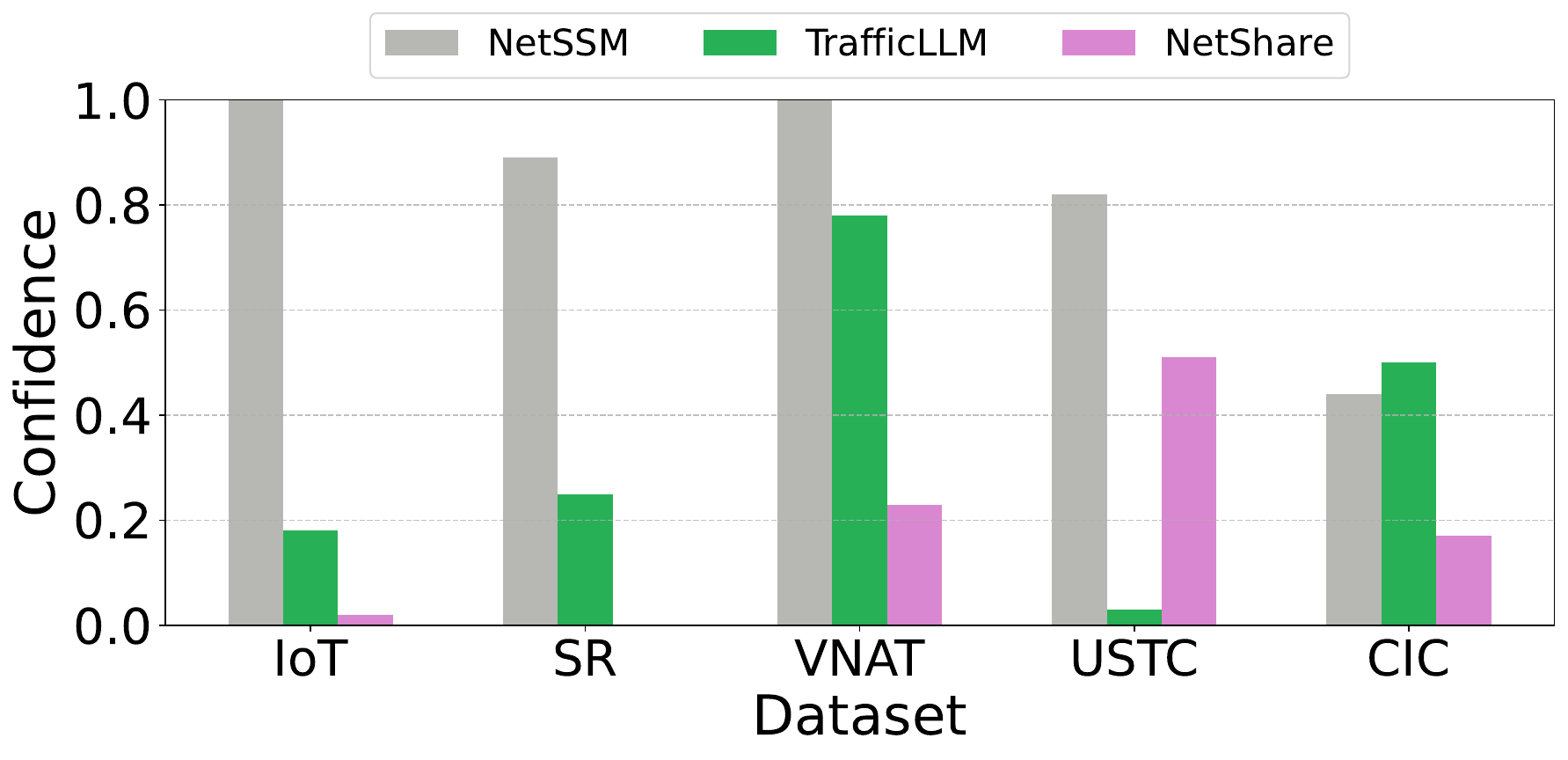}
        \caption{Confidence}
      
    \end{subfigure}
    \hfill
    \begin{subfigure}[b]{\linewidth}
        \centering
        \includegraphics[width=0.75\linewidth]{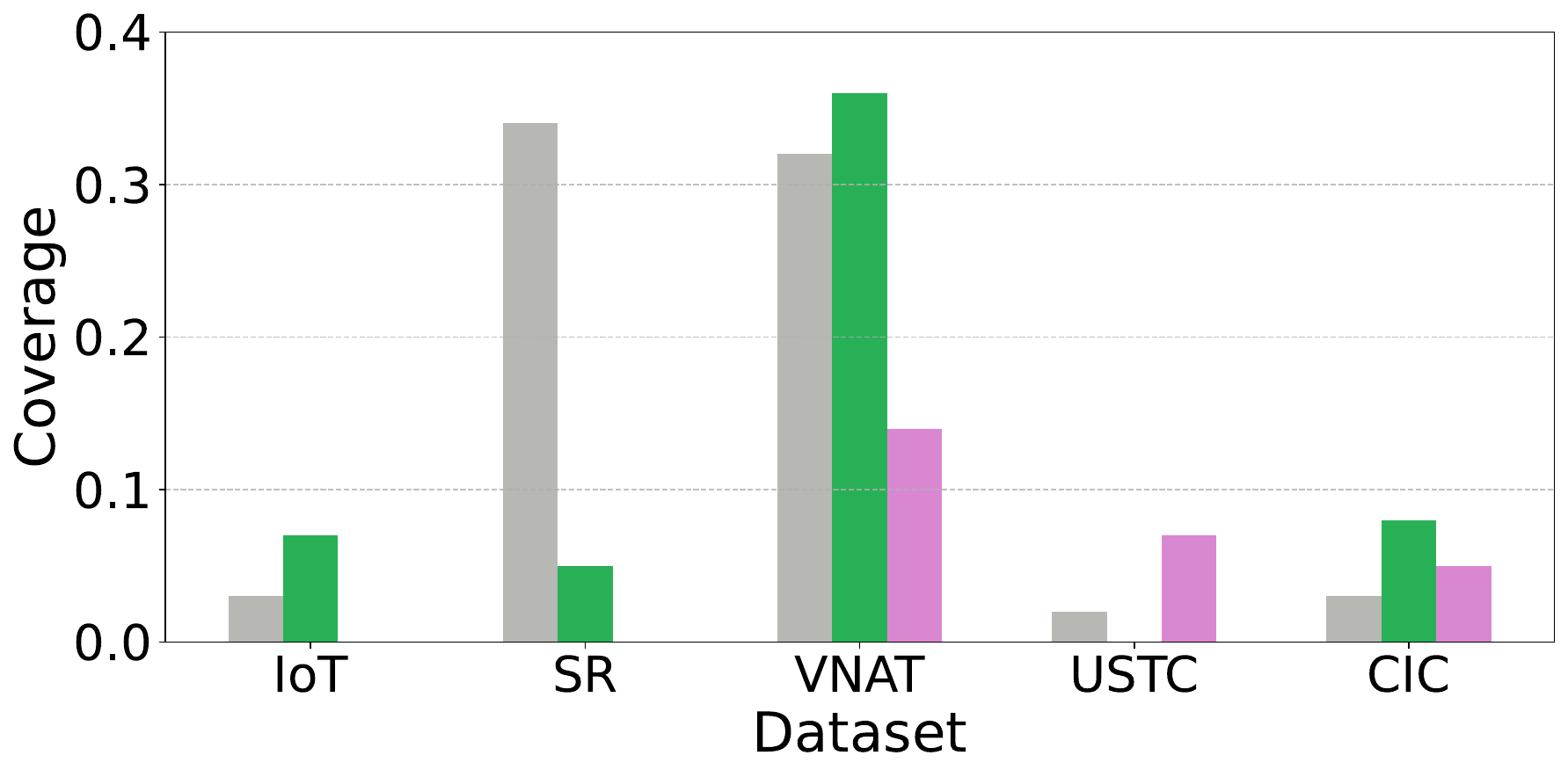}
        \caption{Coverage}
       
    \end{subfigure}

    \caption{Network identifier leakage performance for source IP addresses. We observe high confidence scores across all models, particularly in NetSSM (exceeding 80\% on most datasets), while coverage scores surpass 30\% for NetSSM and TrafficeLLM on two  datasets.}
    \label{fig:leakage-sourceIP}
\end{figure}

\begin{figure}[t]
    \centering
    
  \includegraphics[width=0.75\linewidth]{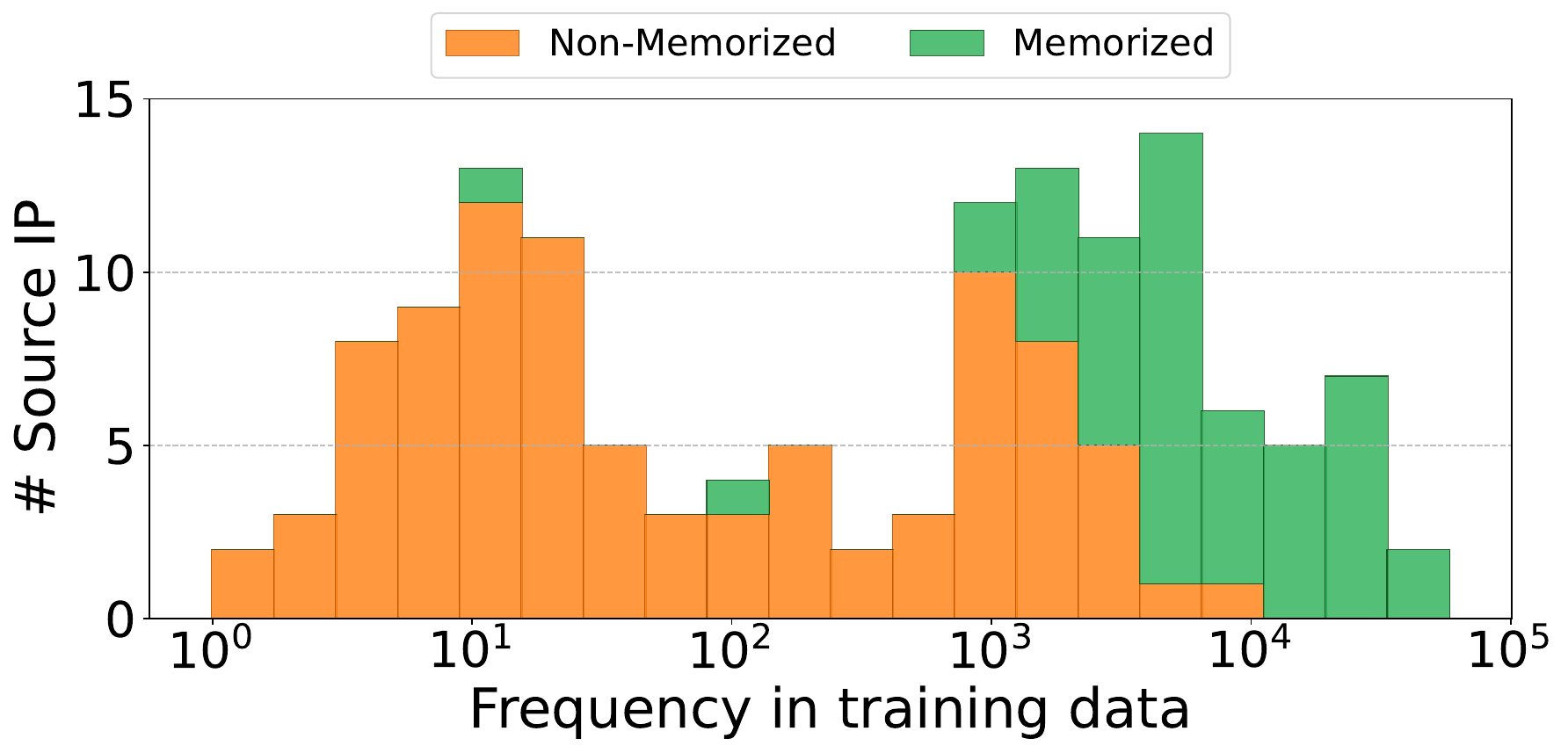}
    \caption{Distribution of memorized and non-memorized source IPs as a function of their frequency in the training data for dataset SR. Source IPs that appear more frequently in the training set are more likely to be memorized and reproduced in the generated output. }\label{fig:memorization-distribution}
\end{figure}

\subsubsection{Network identifiers leakage}

Network identifiers such as source and destination IP addresses exhibit significant leakage in both confidence and coverage metrics (Figure~\ref{fig:leakage-sourceIP} for source IPs; Figure~\ref{fig:leakage-destinationIP} in the Appendix for destination IPs). This leakage is most severe for NetSSM, followed by TrafficLLM and NetShare. For source IPs, confidence scores exceed 0.8 on four of five datasets, while coverage scores reach above 0.35 on some datasets. Given that each dataset contains thousands of unique IPs, these coverage rates indicate exposure of hundreds to thousands of network identifiers.

Furthermore, we observe that identifiers appearing with high frequency in training data are more likely to be memorized (refer to Figure \ref{fig:memorization-distribution} for an example). Specifically, source IPs appearing more than 1,000 times in the training data show substantially higher memorization rates compared to those with lower frequencies. This pattern reveals a critical privacy vulnerability: frequently occurring network identifiers—which often correspond to important servers, major clients, or central network infrastructure—face disproportionately higher exposure risks. This finding has important implications for network privacy: the most prominent and potentially valuable entities in a network are precisely those most vulnerable to memorization.

\subsubsection{Network properties leakage}

\begin{table}[t]
    \centering
    \resizebox{0.47\textwidth}{!}{
    \begin{tabular}{l|rrrr}
    \toprule

      \textbf{Feature} & \textbf{NetDiffusion} &  \textbf{NetSSM} & \textbf{TrafficLLM} & \textbf{NetShare}   \\
    \midrule
      Time to Live (TTL)  &  0.36 & \textbf{0.10}  & 0.22 & 0.32\\
      Identification (ID)  &  0.26 &  \textbf{0.08} & 0.18 & 0.27\\
        Type of Service (ToS)   &  0.17 & \textbf{0.06}  & 0.11 &0.27 \\
        TCP Window Size   &  0.14 &  \textbf{0.12} & 0.25 & -\\
        TCP Control Flags   & \textbf{0.15}  &  \textbf{0.15} & 0.16 & 0.30\\
         TCP Data Offset   & 0.16  & 0.28 & \textbf{0.14} & 0.50\\
        Flow Size   & -  &  \textbf{0.05} & - & 0.13\\
        Packet Size   &  0.14 & \textbf{0.12}  & - & 0.28\\


        \bottomrule
         
    \end{tabular}}
    \caption{Normalized EMD between the original and generated distributions for each sensitive property. Bolded values indicate the lowest distance (more leakage) for each property. }
    \label{tab:divergence}
\end{table}

Without any mitigation methods, the EMD between training and generated data can be as low as 0.10 or less for four key properties (TTL, ID, Flow Size, and ToS), indicating that these features can remain highly useful for fingerprinting attacks. Across all four generative models, NetSSM exhibits the greatest privacy leakage (lowest EMD). For the other three models, leakage varies by feature: some features show substantially higher distances while others remain closely matched to the training distribution. NetShare, a flow-based generative method, exhibits the largest distribution distance overall, suggesting relatively better privacy preservation for sensitive properties compared to packet-based generative models. Refer to Table \ref{tab:divergence} for details.
\begin{figure}[t]
    \centering

  \includegraphics[width=0.75\linewidth]{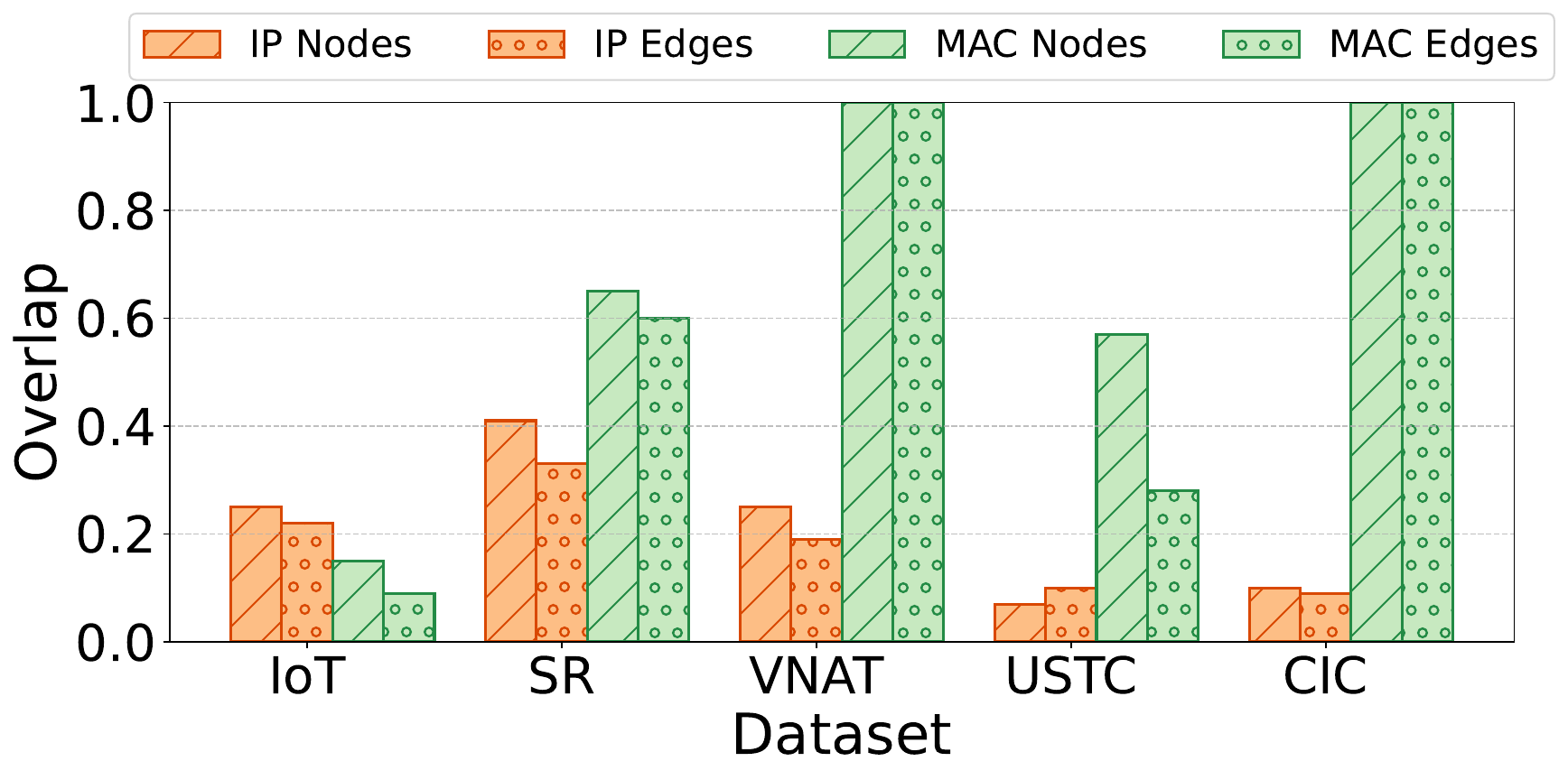}
    \caption{Topology overlap (nodes and edges) for IP-based and MAC-based communications across different datasets. MAC-based communications exhibit higher leakage rates than IP-based communications.}\label{fig:topology}
\end{figure}

\newcommand{\findingfour}[1]{
    \noindent\rule{\columnwidth}{1.5pt}
    
    \noindent\textbf{Finding 4: #1}
    
    \noindent\rule{\columnwidth}{1.5pt}
}

\subsubsection{Network topology leakage}

\begin{table}[t]
    \centering
    \resizebox{0.30\textwidth}{!}{
    \begin{tabular}{l|rr}
    \toprule
        \textbf{Traffic Dataset}&  \multicolumn{2}{c}{\textbf{Degree distribution distance}} \\

        &  \textbf{MAC-based} &  \textbf{IP-based} \\

    \midrule
        IoT &  \textbf{0.04}  &    \textbf{0.01}\\
        SR &   \textbf{0.05}     & \textbf{0.04}\\
        VNAT & 0.17  &   \textbf{0.04}\\
        USTC& 0.19  &    \textbf{0.01}\\
    CIC &  0.40  &    \textbf{0.00}\\

        \bottomrule
         
    \end{tabular}}
    \caption{Normalized EMD comparing node degree distributions in graph topologies constructed from original versus generated data.}
    \label{tab:topo-distance}
\end{table}

NetShare demonstrates minimal network topology leakage with very low node and edge overlap rates. In contrast, NetSSM-generated traffic exhibits substantial topology leakage (see Figure \ref{fig:topology} for details). For IP-based topologies, NetSSM achieves node overlap rates up to 0.4 and edge overlap rates up to 0.35. For MAC-based topologies, overlap rates reach 1.0 for both nodes and edges, indicating complete topology reconstruction. These high overlap rates demonstrate that NetSSM memorizes and reproduces significant portions of the training network topology, including key entities and their communication patterns. The substantially higher MAC-level leakage occurs because MAC address spaces are often smaller than IP address spaces, containing fewer unique addresses and communication flows, making the topology easier to memorize.

\begin{figure}[t]
    \centering
    
  \includegraphics[width=0.84\linewidth]{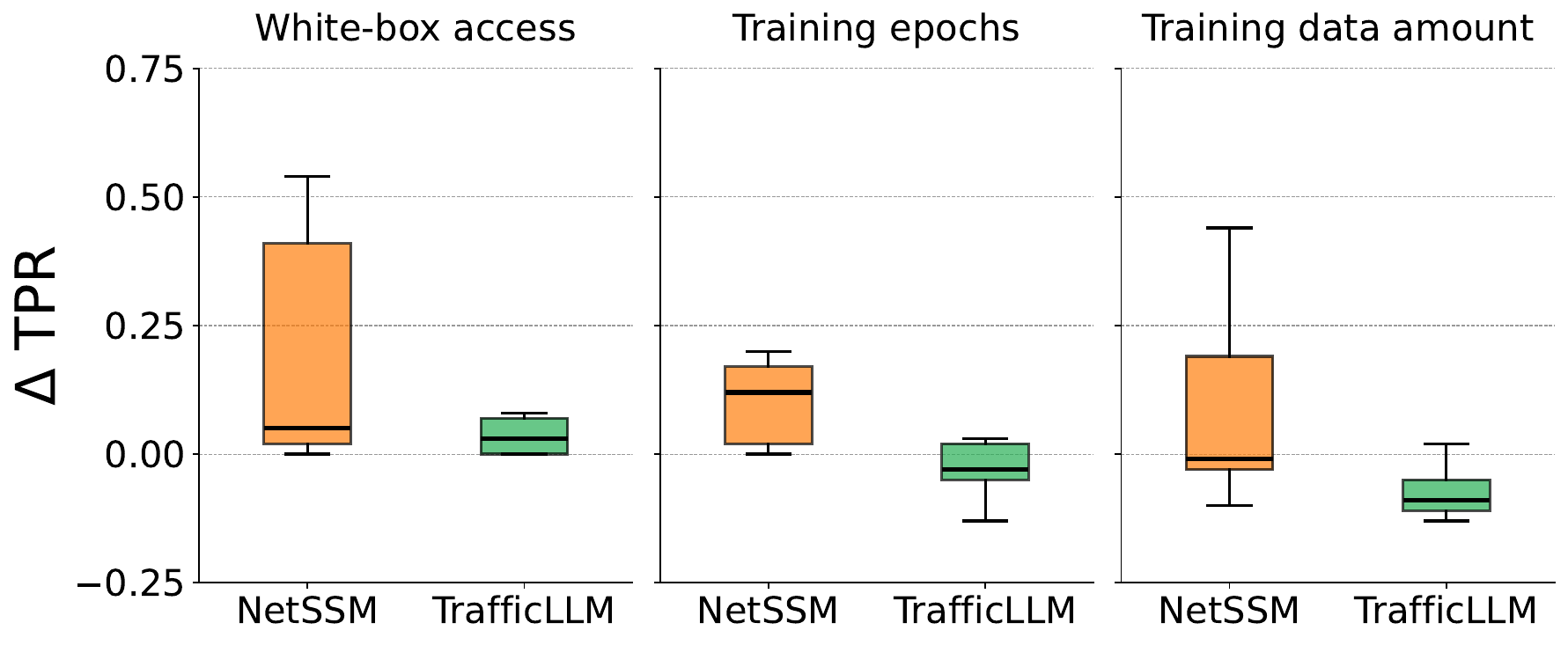}
    \caption{Impact of different experimental conditions on MIA effectiveness under three conditions: (1) white-box versus black-box access, (2) 10 versus 1 epoch of training, and (3) 2x versus 1x training data size.}\label{fig:changes-tpr}
\end{figure}
\begin{figure}[t]
    \centering
    
  \includegraphics[width=0.84\linewidth]{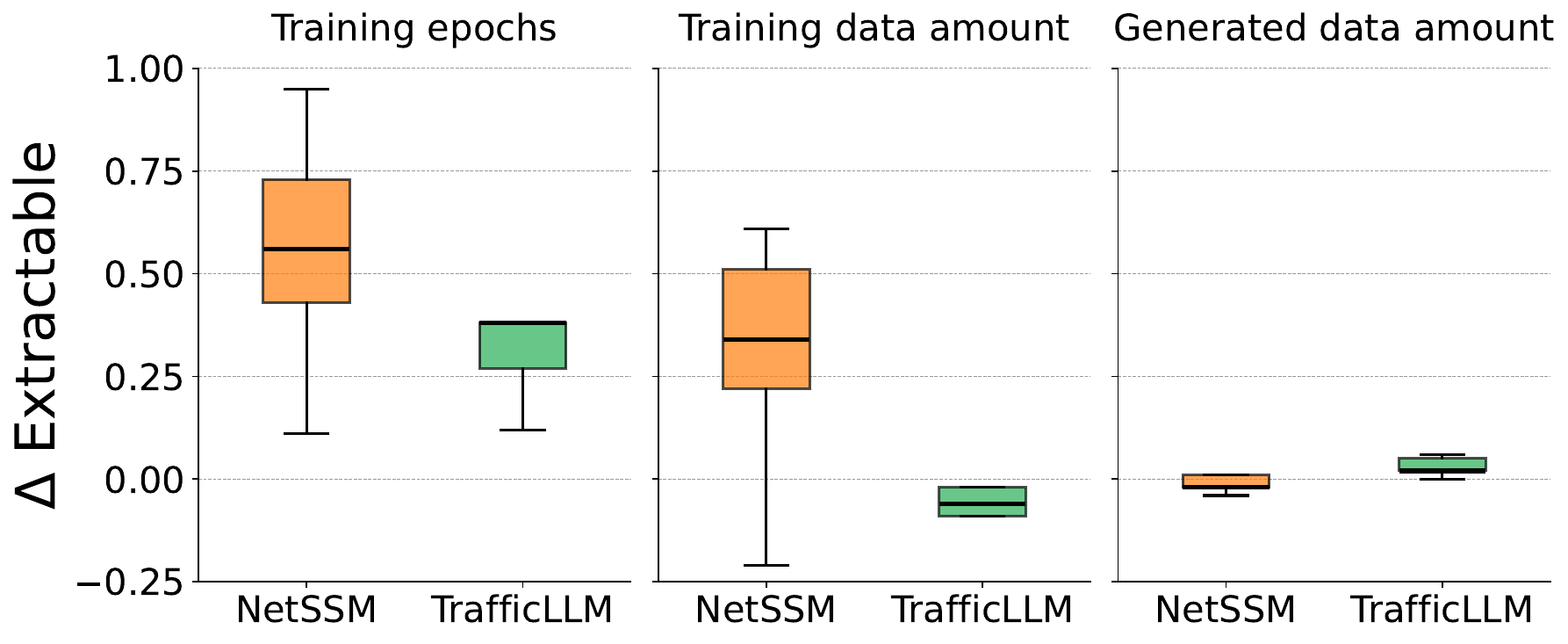}
    \caption{Impact of different experimental conditions on data extraction effectiveness (averaged across datasets). Three conditions are examined: (1) 10 epochs versus 1 epoch of training, (2) 2x versus 1x training data size and (3) 2x versus 1x generated data size.}\label{fig:changes-extractable}
\end{figure}

\findingfour{Critical, high-volume, and highly connected nodes are likely to be exposed in generated topologies.}

Furthermore, the low degree-distribution EMD for overlapping nodes indicates high structural similarity between training and generated topologies. Analysis reveals that these overlapping nodes are predominantly high-degree nodes (hubs) with many connections, likely corresponding to critical network entities such as major routers, central servers, or key communication endpoints. The exposure of these hub nodes and their connectivity patterns reveals the network's structural vulnerabilities and critical infrastructure, which attackers could exploit to identify high-value targets or plan strategic disruptions.

\section{Which factors determine the effectiveness of privacy attacks?}\label{sec:RQ2}

\subsection{Factors Affecting MIA} White-box access and increased training data significantly increase MIA effectiveness, particularly for NetSSM. In contrast, TrafficLLM's MIA vulnerability remains relatively stable or even slightly decreases with these factors. This suggests that for TrafficLLM, additional training epochs and data may dilute membership signals rather than strengthen them. See Figure \ref{fig:changes-tpr} for details on how each factor affects MIA effectiveness across models. 

\vspace{1.95pt}
\newcommand{\findingfive}[1]{
    \noindent\rule{\columnwidth}{1.5pt}
    
    \noindent\textbf{Finding 5: #1}
    
    \noindent\rule{\columnwidth}{1.5pt}
}

\findingfive{White-box access, increased training epochs, and increased training dataset size can significantly increase MIA effectiveness.}

\subsection{Factors Affecting Data Extraction}
As shown in Figure \ref{fig:changes-extractable},
increasing training epochs significantly increases the extractable rate for both NetSSM and TrafficLLM, as more training iterations enhance the model's ability to memorize training data. Increasing the amount of training data also raises the extractable rate for NetSSM, but remain mostly the same for TrafficLLM. Finally, doubling the amount of generated data does not affect the extractable rate, since all generated samples are drawn from the same learned distribution, which remains constant regardless of generation volume. 

\begin{figure}[t]
    \centering

    \begin{subfigure}[b]{0.95\linewidth}
        \centering
        \includegraphics[width=\linewidth]{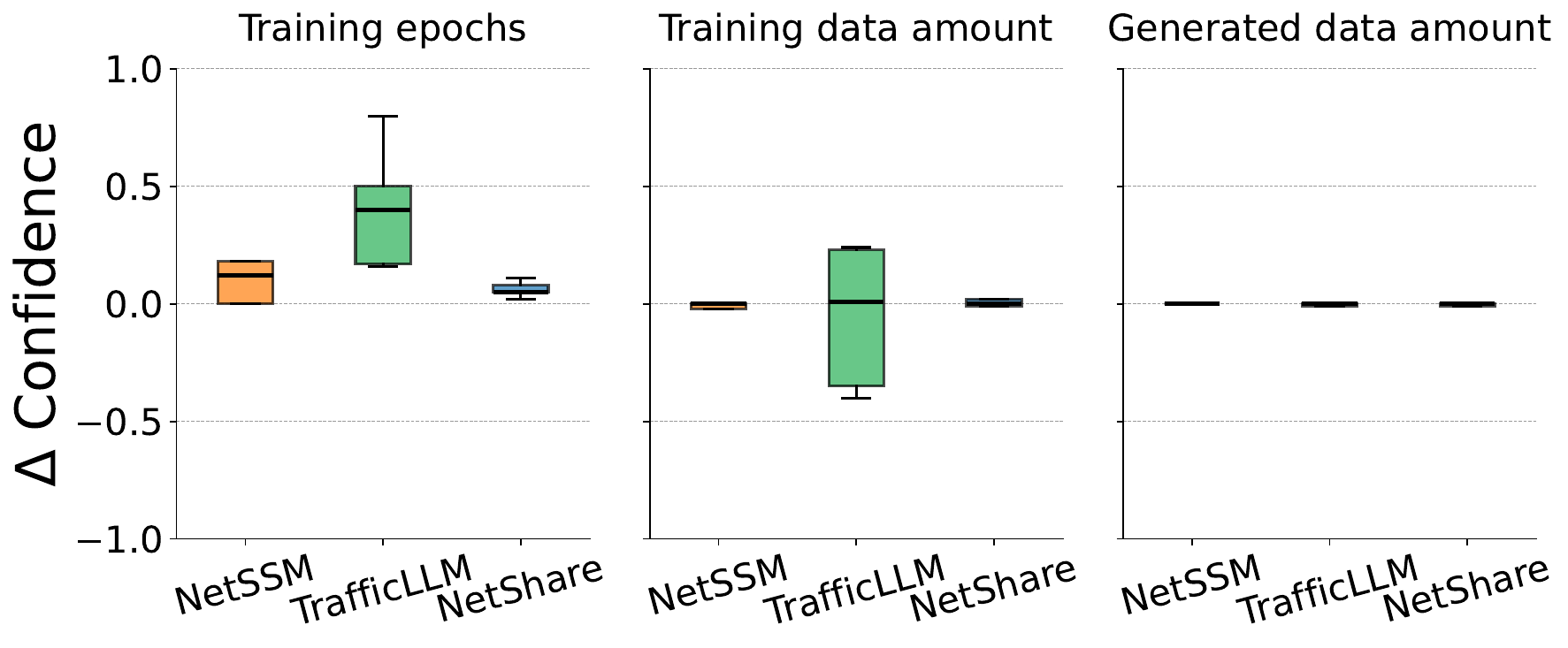}
        \caption{Difference of confidence scores in various settings.}
      
    \end{subfigure}
    \hfill
    \begin{subfigure}[b]{0.95\linewidth}
        \centering
        \includegraphics[width=\linewidth]{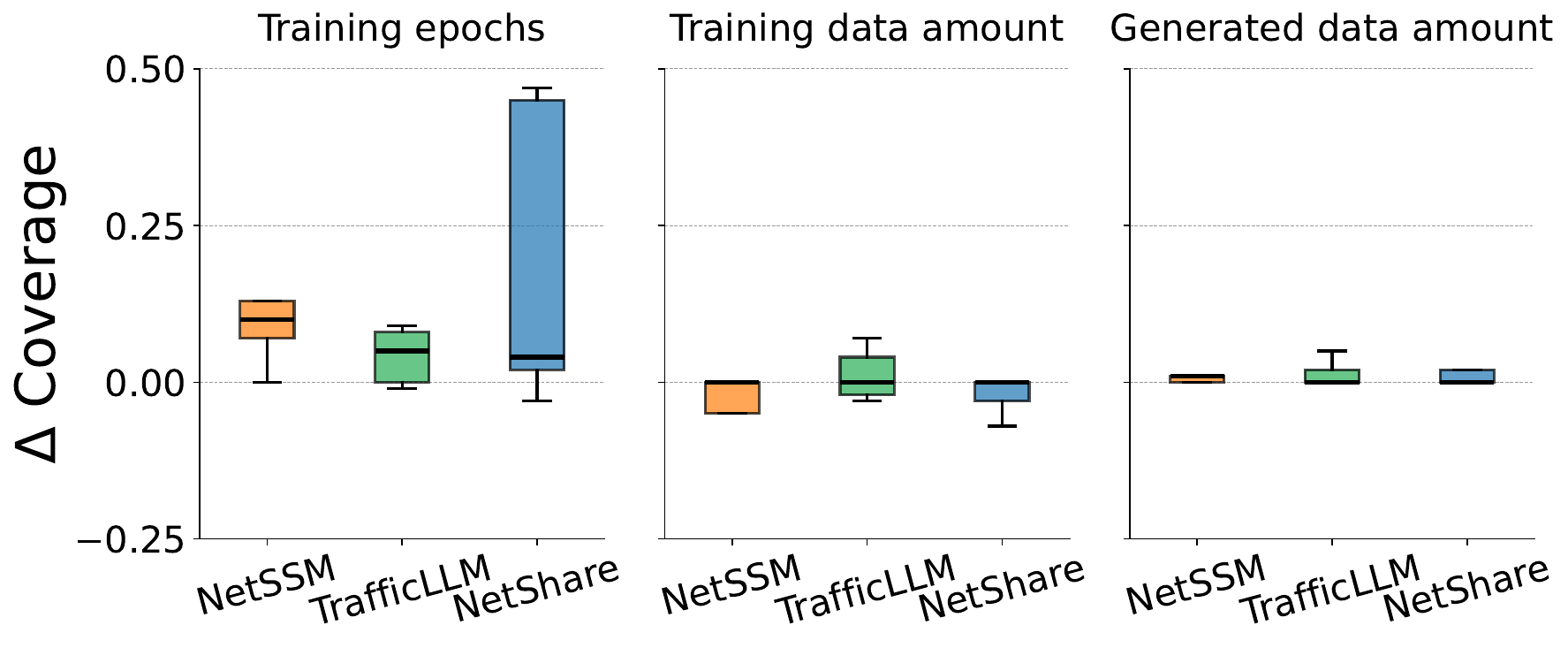}
        \caption{Difference of coverage scores in various settings.}
       
    \end{subfigure}

    \caption{Impact of different experimental conditions on network identifier memorization for source IP (averaged across datasets) under three conditions: (1) 10 versus 1 epoch of training, (2) 2x versus 1x training data size and (3) 2x versus 1x generated data size.}
    \label{fig:leakage-change-sourceIP}
\end{figure}

\subsection{Factors Affecting Network-specific Attacks} Increasing the number of training epochs tends to elevate both confidence and coverage scores. For example, confidence scores rise for NetSSM and TrafficLLM, while coverage scores increase for NetShare. In contrast, doubling the amount of training data yields inconsistent effects on network-specific attacks. For TrafficLLM, confidence scores decrease on some datasets while increasing on others. This mixed pattern arises from two competing effects: while additional training data provides more samples that could be memorized, it also diversifies the training distribution, potentially reducing overfitting. Consequently, confidence scores increase on datasets where memorization dominates but decrease where diversification is stronger. Finally, doubling the amount of generated data does not increase coverage or confidence scores.


\section{What is the privacy-utility tradeoff when applying privacy mitigation techniques?}\label{sec:RQ3}

\begin{table*}[h!]
    \small
    \centering
    \resizebox{0.95\textwidth}{!}{
    \begin{tabular}{lll|rrr|rr} 
    \toprule
    \textbf{Generation}&\textbf{Mitigation}& \textbf{Detail} & \multicolumn{3}{c|}{\textbf{ Privacy attack effectiveness}}& \multicolumn{2}{c}{\textbf{Data Utility}}  \\
       \textbf{task}& \textbf{strategy}&  & \textbf{MIA} & \textbf{Extraction}& \textbf{Network} & \textbf{Fidelity} & \textbf{Task accuracy} \\
       & &  & (\textcolor{darkgreen}{$\downarrow$} better)
 &(\textcolor{darkgreen}{$\downarrow$} better) & (\textcolor{darkgreen}{$\downarrow$} better)  & (\textcolor{darkgreen}{$\uparrow$} better) & (\textcolor{darkgreen}{$\uparrow$} better) \\
    \midrule
     Multi-flow & Anonymization & Complete anonymization (CA) & \textcolor{darkgreen}{$\downarrow$} -0.03&\textcolor{darkgreen}{$\downarrow$}  \textbf{-0.43}& 0.02 & -0.15 &-\\
       & &Pseudonymization (PS) & \textcolor{darkgreen}{$\downarrow$}\textbf{-0.18}& \textcolor{darkgreen}{$\downarrow$} \textbf{-0.43}& 0.00 &-0.14 &- \\
        & &Privacy-preserving (PP) & +0.02& \textcolor{darkgreen}{$\downarrow$} \textbf{-0.43} &0.03 &  -0.06 &-\\
         
        &DP-Noise  &Epsilon = 0.1 & 0.07 & \textcolor{darkgreen}{$\downarrow$}-0.08&\textcolor{darkgreen}{$\downarrow$}\textbf{-0.11} &\textbf{-0.03} &- \\
        & &Epsilon = 1 &0.02 &0.04 &\textcolor{darkgreen}{$\downarrow$}-0.08 &\textbf{-0.03} &- \\
    \midrule
     Single-flow & Anonymization & Complete anonymization (CA) & \textcolor{darkgreen}{$\downarrow$}\textbf{-0.38}& \textcolor{darkgreen}{$\downarrow$}\textbf{-0.91}&0.11 & -0.08& -0.11\\
       & &Pseudonymization (PS) & \textcolor{darkgreen}{$\downarrow$}-0.14& \textcolor{darkgreen}{$\downarrow$}\textbf{-0.91}&0.02 &-0.05& -0.05 \\
        & &Privacy-preserving (PP) & \textcolor{darkgreen}{$\downarrow$}-0.23& \textcolor{darkgreen}{$\downarrow$}\textbf{-0.91} & 0.04& 0.00& -0.04\\
         
        &DP-Noise  &Epsilon = 0.1 &\textcolor{darkgreen}{$\downarrow$}-0.03 & 0.00& \textcolor{darkgreen}{$\downarrow$}\textbf{-0.07}& -0.02 &\textcolor{darkgreen}{$\uparrow$}\textbf{0.02}\\
        & &Epsilon = 1 &0.01 & 0.00& \textcolor{darkgreen}{$\downarrow$}-0.05&\textcolor{darkgreen}{$\uparrow$}\textbf{0.01} & -0.01\\

    \bottomrule   
    \end{tabular}
    }
    \caption{Changes in privacy and utility metrics when applying mitigation techniques to training data. \textcolor{darkgreen}{$\downarrow$} indicates improved privacy protection (attack becomes less effective); \textcolor{darkgreen}{$\uparrow$} indicates improved synthetic data utility.
Bold values indicate the best performance for each metric. Note that downstream task accuracy for multi-flow is not evaluated, because none of the datasets have labels for multi-flow traffic.}
    \label{tab:mit-results}
\end{table*}

To address RQ3, we evaluate how privacy mitigation strategies influence both privacy risks and downstream utility. We focus on NetSSM because it exhibits the strongest leakage across the attacks studied in Sections~5 and~6, and we evaluate both its multi-flow and single-flow generation settings. As stated in Section~\ref{sec:method}, we utilize both anonymization and noise-based mitigation. For anonymization, we study three strategies: \emph{Complete Anonymization} (CA), \emph{Pseudonymization} (PS), and \emph{Privacy-preserving} (PP). Privacy-preserving remove low-bits from the each identifier, thereby hiding host-level identifiers while preserving coarse structural information. For noise-based mitigation, we add DP noise to sensitive header fields using privacy parameters $\epsilon=0.1$ and $\epsilon=1$, then apply clipping and rounding to keep values valid. 

We quantify privacy and utility changes relative to the baseline without mitigation using: (i) MIA TPR@FPR$\leq 0.01$, (ii) extractable rate in data extraction, (iii) sensitive properties (the decrease in EMD means an increase in privacy attack effectiveness), (iv) fidelity based on EMD (the decrease in EMD means an increase in fidelity) and (v) downstream classification accuracy for IoT and SR (single-flow generation only). Table~\ref{tab:mit-results} summarizes the results.

\newcommand{\findingsix}[1]{
    \noindent\rule{\columnwidth}{1.5pt}
    
    \noindent\textbf{Finding 6: #1}
    
    \noindent\rule{\columnwidth}{1.5pt}
}

\findingsix{Anonymization reduces extraction and membership signals, with moderate utility loss.}

Across both multi-flow and single-flow settings, all three anonymization strategies eliminate data extraction risks. When applied to training data, CA, PS, and PP reduce the extractable rate to 0. This indicates that header-level memorization is strongly tied to raw identifiers; once identifiers are removed or remapped, memorized subsequences almost never reappear in generated outputs. Anonymization also lowers MIA TPR, especially in the single-flow setting: CA and PP reduce TPR by $0.38$ and $0.23$ respectively, while PS yields a smaller reduction of $0.14$. In the multi-flow setting, PS is the most effective, while CA and PP have smaller effects.

Anonymization, however, reduces utility. For single-flow generation, IoT and SR classification accuracy declines by up to $0.11$ for CA and by $0.04$--$0.05$ for PS and PP. Fidelity shifts are smaller than the changes observed for MIA or extraction, suggesting that NetSSM can still approximate the anonymized distribution reasonably well. Network properties inference effectiveness may still increase after anonymization when evaluated over anonymized identifiers, showing the improvement in one privacy attack type does not necessarily correlates with other privacy attacks.

\newcommand{\findingseven}[1]{
    \noindent\rule{\columnwidth}{1.5pt}
    
    \noindent\textbf{Finding 7: #1}
    
    \noindent\rule{\columnwidth}{1.5pt}
}
\findingseven{DP noise protects against network properties inference attack while preserving utility.}

Adding DP noise ($\epsilon=0.1$ or $\epsilon=1$) produces a distinct tradeoff. Extractable rate decreases slightly in multi-flow generation (approximately $0.08$ for $\epsilon=0.1$) or remains unchanged in single-flow settings, consistent with noise perturbing marginal distributions rather than token sequences. Changes in MIA TPR are small and occasionally positive, showing that noise on a few header fields does not reliably weaken membership signals for sequence models.

In contrast, DP noise consistently suppresses leakage of network properties. Network properties inference effectiveness decreases by $0.05$--$0.11$ depending on $\epsilon$ and dataset, because DP perturbation is directly added to protect these properties. Utility remains largely intact: classification accuracy differs by at most $\pm 0.02$, with a slight improvement for $\epsilon=0.1$, and fidelity shifts are minimal. These results indicate that moderate perturbations to header statistics weaken fingerprinting-style leakage while keeping downstream utility nearly unchanged.

\textbf{Multi-flow vs. Single-flow Differences.}
Mitigation has a larger impact on privacy in single-flow settings. Single-flow generation amplifies per-entity regularities, resulting in stronger leakage and proportionally stronger mitigation gains. However, the single-flow setting is also the only setting used for downstream flow classification, and it is where anonymization reduces task accuracy. Multi-flow generation focuses on aggregate cross-flow behavior, so anonymization and DP noise modify identifiers and subtle header distributions but leave coarse traffic structure largely unaffected.




\section{Discussion} \label{sec:discussion}
\textbf{The Dual Nature of Privacy and Fidelity. }A fundamental challenge in generating synthetic network traffic lies in the complex and often paradoxical relationship between privacy and fidelity. Many metrics that serve as indicators of high-fidelity synthetic data can simultaneously constitute significant privacy risks. For instance, accurately reproducing flow duration distributions, packet inter-arrival times, or port usage patterns may be essential for downstream tasks such as intrusion detection or QoS analysis but can also be exploited by adversaries to infer sensitive information about network topology and user behaviors.
This duality means that improving one objective often degrades the other, creating a fundamental tension that cannot be resolved through technical means alone but requires careful consideration of use case requirements and threat models.

\textbf{Beyond Membership Inference.} Our findings also reveal privacy concerns for network traffic extended beyond MIA. Some datasets (such as USTC and CIC) may not be vulnerable to MIA attacks (refer to Figure \ref{fig:mia_basecase}) but are still vulnerable to other privacy attacks such as data extraction (Figure \ref{fig:extractable-position-NetSSM}) or identifier memorization (Figure \ref{fig:leakage-sourceIP}). Hence, characterizing and quantifying these diverse privacy risks is critical for developing a comprehensive understanding of the privacy implications of synthetic network traffic. A sole focus on MIA can create a false sense of security, as synthetic data may perform well against membership inference while remaining vulnerable to other forms of privacy breaches that could be equally or more damaging in practice.

\textbf{Context-Aware Privacy-Utility Tradeoff.} There is no universal solution to the privacy-utility tradeoff in synthetic network traffic generation; the appropriate balance depends on the specific use case, threat model, and sharing context. For internal use within trusted environments, fine-grained generative models such as NetSSM can prioritize fidelity, retaining rich packet-level features with minimal anonymization to support comprehensive analysis. In contrast, external or public data sharing requires stronger protections, including but not limited to using less fine-granular methods such as NetDiffusion and NetShare, anonymization of critical infrastructure and adding differential privacy noise. However, it ultimately comes with some degradations of utility. In general, privacy and utility are context-dependent, and effective mitigation demands aligning protection levels with intended use and potential adversarial capabilities.

\section{Conclusion} \label{sec:conclusion}

This paper presents a systematic study of privacy risks in synthetic network traffic using membership inference, data extraction, and network-specific metrics for identifiers, attributes, and topology. Across four state-of-the-art generators and five datasets, we showed that synthetic traces can leak training membership, memorize header tokens, and reproduce a substantial fraction of network identifiers and communication structure. We further analyzed how attacker access, training duration, data scale, and model overfitting shape these risks, and we evaluated anonymization- and noise-based mitigations, quantifying their privacy–utility tradeoffs.
Overall, our results indicate that synthetic network traces should be treated as security-sensitive data rather than as inherently safe substitutes for raw traffic. Practitioners who deploy such models should evaluate privacy along multiple dimensions and accept losses in fidelity when stronger protection is required. Future work includes extending to design stronger privacy-enhancing methods, and building benchmarks that jointly report fidelity, downstream utility, and privacy risk to guide method selection and deployment.

\bibliographystyle{IEEEtran}
\bibliography{ref}

@inproceedings{lacage2006yet,
  title={Yet another network simulator},
  author={Lacage, Mathieu and Henderson, Thomas R},
  booktitle={Proceedings of the 2006 Workshop on ns-3},
  pages={12--es},
  year={2006}
}

@article{henderson2008network,
  title={Network simulations with the ns-3 simulator},
  author={Henderson, Thomas R and Lacage, Mathieu and Riley, George F and Dowell, Craig and Kopena, Joseph},
  journal={SIGCOMM demonstration},
  volume={14},
  number={14},
  pages={527},
  year={2008}
}

@misc{NetSimTM,
  author       = {Boson},
  title        = {NetSimTM Network SimulatorTM \& Router Simulator},
  year         = {n.d},
  howpublished = {\url{https://www.boson.com/netsim-cisco-network-simulator}},
  note         = {Accessed Oct 2, 2025}
}

@misc{OMNeT,
  author       = {OMNeT++},
  title        = {OMNeT Discreet Event Simulator},
  year         = {n.d},
  howpublished = {\url{https://omnetpp.org/}},
  note         = {Accessed Oct 2, 2025}
}

@inproceedings{zeng1998glomosim,
  title={GloMoSim: a library for parallel simulation of large-scale wireless networks},
  author={Zeng, Xiang and Bagrodia, Rajive and Gerla, Mario},
  booktitle={Proceedings of the twelfth workshop on Parallel and distributed simulation},
  pages={154--161},
  year={1998}
}

@inproceedings{riley2003large,
  title={Large-scale network simulations with GTNetS},
  author={Riley},
  booktitle={Proceedings of the 2003 Winter Simulation Conference, 2003.},
  volume={1},
  pages={676--684},
  year={2003},
  organization={IEEE}
}

@inproceedings{buss1996discrete,
  title={Discrete-event simulation on the world wide web using Java},
  author={Buss, Arnold H and Stork, Kirk A},
  booktitle={Proceedings of the 28th conference on Winter Simulation},
  pages={780--785},
  year={1996}
}

@article{redvzovic2017ip,
  title={IP traffic generator based on hidden Markov models},
  author={Red{\v{z}}ovi{\'c}, Hasan and Smiljani{\'c}, Aleksandra and Bjelica, Milan},
  journal={parameters},
  volume={1},
  number={2},
  pages={1},
  year={2017}
}

@inproceedings{xu2021stan,
  title={Stan: Synthetic network traffic generation with generative neural models},
  author={Xu, Shengzhe and Marwah, Manish and Arlitt, Martin and Ramakrishnan, Naren},
  booktitle={International Workshop on Deployable Machine Learning for Security Defense},
  pages={3--29},
  year={2021},
  organization={Springer}
}

@inproceedings{sun2024netdpsyn,
  title={Netdpsyn: synthesizing network traces under differential privacy},
  author={Sun, Danyu and Chen, Joann Qiongna and Gong, Chen and Wang, Tianhao and Li, Zhou},
  booktitle={Proceedings of the 2024 ACM on Internet Measurement Conference},
  pages={545--554},
  year={2024}
}

@inproceedings{lin2020using,
  title={Using gans for sharing networked time series data: Challenges, initial promise, and open questions},
  author={Lin, Zinan and Jain, Alankar and Wang, Chen and Fanti, Giulia and Sekar, Vyas},
  booktitle={Proceedings of the ACM internet measurement conference},
  pages={464--483},
  year={2020}
}

@inproceedings{wang2020packetcgan,
  title={PacketCGAN: Exploratory study of class imbalance for encrypted traffic classification using CGAN},
  author={Wang, Pan and Li, Shuhang and Ye, Feng and Wang, Zixuan and Zhang, Moxuan},
  booktitle={ICC 2020-2020 IEEE International Conference on Communications (ICC)},
  pages={1--7},
  year={2020},
  organization={IEEE}
}

@inproceedings{yin2022practical,
  title={Practical gan-based synthetic ip header trace generation using netshare},
  author={Yin, Yucheng and Lin, Zinan and Jin, Minhao and Fanti, Giulia and Sekar, Vyas},
  booktitle={Proceedings of the ACM SIGCOMM 2022 Conference},
  pages={458--472},
  year={2022}
}

@article{jiang2024netdiffusion,
  title={Netdiffusion: Network data augmentation through protocol-constrained traffic generation},
  author={Jiang, Xi and Liu, Shinan and Gember-Jacobson, Aaron and Bhagoji, Arjun Nitin and Schmitt, Paul and Bronzino, Francesco and Feamster, Nick},
  journal={Proceedings of the ACM on Measurement and Analysis of Computing Systems},
  volume={8},
  number={1},
  pages={1--32},
  year={2024},
  publisher={ACM New York, NY, USA}
}

@article{zhang2024netdiff,
  title={NetDiff: a service-guided hierarchical diffusion model for network flow trace generation},
  author={Zhang, Shiyuan and Li, Tong and Jin, Depeng and Li, Yong},
  journal={Proceedings of the ACM on Networking},
  volume={2},
  number={CoNEXT3},
  pages={1--21},
  year={2024},
  publisher={ACM New York, NY, USA}
}

@ARTICLE{GDPlan,
  author={Kan, Nuowen and Yan, Sa and Zou, Junni and Dai, Wenrui and Gao, Xing and Li, Chenglin and Xiong, Hongkai},
  journal={IEEE Transactions on Networking}, 
  title={GDPlan: Generative Network Planning via Graph Diffusion Model}, 
  year={2025},
  volume={33},
  number={4},
  pages={1422-1437},
  doi={10.1109/TON.2025.3535518}}

@article{meng2023netgpt,
  title={Netgpt: Generative pretrained transformer for network traffic},
  author={Meng, Xuying and Lin, Chungang and Wang, Yequan and Zhang, Yujun},
  journal={arXiv preprint arXiv:2304.09513},
  year={2023}
}

@article{qu2024trafficgpt,
  title={Trafficgpt: Breaking the token barrier for efficient long traffic analysis and generation},
  author={Qu, Jian and Ma, Xiaobo and Li, Jianfeng},
  journal={arXiv preprint arXiv:2403.05822},
  year={2024}
}

@article{chu2025netssm,
  title={NetSSM: Multi-Flow and State-Aware Network Trace Generation using State-Space Models},
  author={Chu, Andrew and Jiang, Xi and Liu, Shinan and Bhagoji, Arjun and Bronzino, Francesco and Schmitt, Paul and Feamster, Nick},
  journal={arXiv preprint arXiv:2503.22663},
  year={2025}
}

@article{sivaroopan2024netdiffus,
  title={Netdiffus: Network traffic generation by diffusion models through time-series imaging},
  author={Sivaroopan, Nirhoshan and Bandara, Dumindu and Madarasingha, Chamara and Jourjon, Guillaume and Jayasumana, Anura P and Thilakarathna, Kanchana},
  journal={Computer Networks},
  volume={251},
  pages={110616},
  year={2024},
  publisher={Elsevier}
}

@article{cui2025trafficllm,
  title={Trafficllm: Enhancing large language models for network traffic analysis with generic traffic representation},
  author={Cui, Tianyu and Lin, Xinjie and Li, Sijia and Chen, Miao and Yin, Qilei and Li, Qi and Xu, Ke},
  journal={arXiv preprint arXiv:2504.04222},
  year={2025}
}

@inproceedings{duan2023diffusion,
  title={Are diffusion models vulnerable to membership inference attacks?},
  author={Duan, Jinhao and Kong, Fei and Wang, Shiqi and Shi, Xiaoshuang and Xu, Kaidi},
  booktitle={International Conference on Machine Learning},
  pages={8717--8730},
  year={2023},
  organization={PMLR}
}

@article{pang2023black,
  title={Black-box membership inference attacks against fine-tuned diffusion models},
  author={Pang, Yan and Wang, Tianhao},
  journal={arXiv preprint arXiv:2312.08207},
  year={2023}
}

@article{hu2023membership,
  title={Membership inference of diffusion models},
  author={Hu, Hailong and Pang, Jun},
  journal={arXiv preprint arXiv:2301.09956},
  year={2023}
}

@inproceedings{matsumoto2023membership,
  title={Membership inference attacks against diffusion models},
  author={Matsumoto, Tomoya and Miura, Takayuki and Yanai, Naoto},
  booktitle={2023 IEEE Security and Privacy Workshops (SPW)},
  pages={77--83},
  year={2023},
  organization={IEEE}
}

@inproceedings{carlini2023extracting,
  title={Extracting training data from diffusion models},
  author={Carlini, Nicolas and Hayes, Jamie and Nasr, Milad and Jagielski, Matthew and Sehwag, Vikash and Tramer, Florian and Balle, Borja and Ippolito, Daphne and Wallace, Eric},
  booktitle={32nd USENIX security symposium (USENIX Security 23)},
  pages={5253--5270},
  year={2023}
}

@article{duan2024membership,
  title={Do membership inference attacks work on large language models?},
  author={Duan, Michael and Suri, Anshuman and Mireshghallah, Niloofar and Min, Sewon and Shi, Weijia and Zettlemoyer, Luke and Tsvetkov, Yulia and Choi, Yejin and Evans, David and Hajishirzi, Hannaneh},
  journal={arXiv preprint arXiv:2402.07841},
  year={2024}
}

@inproceedings{song2025mias,
  title={Em-mias: Enhancing membership inference attacks in large language models through ensemble modeling},
  author={Song, Zichen and Huang, Sitan and Kang, Zhongfeng},
  booktitle={ICASSP 2025-2025 IEEE International Conference on Acoustics, Speech and Signal Processing (ICASSP)},
  pages={1--5},
  year={2025},
  organization={IEEE}
}

@article{mozaffari2024semantic,
  title={Semantic membership inference attack against large language models},
  author={Mozaffari, Hamid and Marathe, Virendra J},
  journal={arXiv preprint arXiv:2406.10218},
  year={2024}
}

@article{ran2025lora,
  title={LoRA-Leak: Membership Inference Attacks Against LoRA Fine-tuned Language Models},
  author={Ran, Delong and He, Xinlei and Cong, Tianshuo and Wang, Anyu and Li, Qi and Wang, Xiaoyun},
  journal={arXiv preprint arXiv:2507.18302},
  year={2025}
}

@article{jagannatha2021membership,
  title={Membership inference attack susceptibility of clinical language models},
  author={Jagannatha, Abhyuday and Rawat, Bhanu Pratap Singh and Yu, Hong},
  journal={arXiv preprint arXiv:2104.08305},
  year={2021}
}

@INPROCEEDINGS{8429311,
  author={Yeom, Samuel and Giacomelli, Irene and Fredrikson, Matt and Jha, Somesh},
  booktitle={2018 IEEE 31st Computer Security Foundations Symposium (CSF)}, 
  title={Privacy Risk in Machine Learning: Analyzing the Connection to Overfitting}, 
  year={2018},
  volume={},
  number={},
  pages={268-282},
  doi={10.1109/CSF.2018.00027}}

@inproceedings{carlini2021extracting,
  title={Extracting training data from large language models},
  author={Carlini, Nicholas and Tramer, Florian and Wallace, Eric and Jagielski, Matthew and Herbert-Voss, Ariel and Lee, Katherine and Roberts, Adam and Brown, Tom and Song, Dawn and Erlingsson, Ulfar and others},
  booktitle={30th USENIX security symposium (USENIX Security 21)},
  pages={2633--2650},
  year={2021}
}

@inproceedings {299573,
author = {Matthieu Meeus and Shubham Jain and Marek Rei and Yves-Alexandre de Montjoye},
title = {Did the Neurons Read your Book? Document-level Membership Inference for Large Language Models},
booktitle = {33rd USENIX Security Symposium (USENIX Security 24)},
year = {2024},
isbn = {978-1-939133-44-1},
address = {Philadelphia, PA},
pages = {2369--2385},
url = {https://www.usenix.org/conference/usenixsecurity24/presentation/meeus},
publisher = {USENIX Association},
month = aug
}

@article{mattern2023membership,
  title={Membership inference attacks against language models via neighbourhood comparison},
  author={Mattern, Justus and Mireshghallah, Fatemehsadat and Jin, Zhijing and Sch{\"o}lkopf, Bernhard and Sachan, Mrinmaya and Berg-Kirkpatrick, Taylor},
  journal={arXiv preprint arXiv:2305.18462},
  year={2023}
}

@inproceedings{ren2025self,
  title={Self-comparison for dataset-level membership inference in large (vision-) language model},
  author={Ren, Jie and Chen, Kangrui and Chen, Chen and Sehwag, Vikash and Xing, Yue and Tang, Jiliang and Lyu, Lingjuan},
  booktitle={Proceedings of the ACM on Web Conference 2025},
  pages={910--920},
  year={2025}
}

@article{nasr2023scalable,
  title={Scalable extraction of training data from (production) language models},
  author={Nasr, Milad and Carlini, Nicholas and Hayase, Jonathan and Jagielski, Matthew and Cooper, A Feder and Ippolito, Daphne and Choquette-Choo, Christopher A and Wallace, Eric and Tram{\`e}r, Florian and Lee, Katherine},
  journal={arXiv preprint arXiv:2311.17035},
  year={2023}
}

@InProceedings{Tinsley_2021_WACV,
    author    = {Tinsley, Patrick and Czajka, Adam and Flynn, Patrick},
    title     = {This Face Does Not Exist... But It Might Be Yours! Identity Leakage in Generative Models},
    booktitle = {Proceedings of the IEEE/CVF Winter Conference on Applications of Computer Vision (WACV)},
    month     = {January},
    year      = {2021},
    pages     = {1320-1328}
}

@article{liu2024unstoppable,
  title={Unstoppable attack: Label-only model inversion via conditional diffusion model},
  author={Liu, Rongke and Wang, Dong and Ren, Yizhi and Wang, Zhen and Guo, Kaitian and Qin, Qianqian and Liu, Xiaolei},
  journal={IEEE Transactions on Information Forensics and Security},
  volume={19},
  pages={3958--3973},
  year={2024},
  publisher={IEEE}
}

@article{yan2024protecting,
  title={On protecting the data privacy of large language models (llms): A survey},
  author={Yan, Biwei and Li, Kun and Xu, Minghui and Dong, Yueyan and Zhang, Yue and Ren, Zhaochun and Cheng, Xiuzhen},
  journal={arXiv preprint arXiv:2403.05156},
  year={2024}
}

@article{zhou2021property,
  title={Property inference attacks against GANs},
  author={Zhou, Junhao and Chen, Yufei and Shen, Chao and Zhang, Yang},
  journal={arXiv preprint arXiv:2111.07608},
  year={2021}
}

@inproceedings{zhang2020secret,
  title={The secret revealer: Generative model-inversion attacks against deep neural networks},
  author={Zhang, Yuheng and Jia, Ruoxi and Pei, Hengzhi and Wang, Wenxiao and Li, Bo and Song, Dawn},
  booktitle={Proceedings of the IEEE/CVF conference on computer vision and pattern recognition},
  pages={253--261},
  year={2020}
}

@article{hayes2017logan,
  title={Logan: Membership inference attacks against generative models},
  author={Hayes, Jamie and Melis, Luca and Danezis, George and De Cristofaro, Emiliano},
  journal={arXiv preprint arXiv:1705.07663},
  year={2017}
}

@article{hu2021model,
  title={Model extraction and defenses on generative adversarial networks},
  author={Hu, Hailong and Pang, Jun},
  journal={arXiv preprint arXiv:2101.02069},
  year={2021}
}

@article{jin2025assessing,
  title={Assessing User Privacy Leakage in Synthetic Packet Traces: An Attack-Grounded Approach},
  author={Jin, Minhao and He, Hongyu and Apostolaki, Maria},
  journal={arXiv preprint arXiv:2508.11742},
  year={2025}
}

@inproceedings{shokri2017membership,
  title={Membership inference attacks against machine learning models},
  author={Shokri, Reza and Stronati, Marco and Song, Congzheng and Shmatikov, Vitaly},
  booktitle={2017 IEEE symposium on security and privacy (SP)},
  pages={3--18},
  year={2017},
  organization={IEEE}
}

@article{hilprecht2019monte,
  title={Monte carlo and reconstruction membership inference attacks against generative models},
  author={Hilprecht, Benjamin and H{\"a}rterich, Martin and Bernau, Daniel},
  journal={Proceedings on Privacy Enhancing Technologies},
  year={2019}
}

@article{liu2023amir,
  title={Amir: Active multimodal interaction recognition from video and network traffic in connected environments},
  author={Liu, Shinan and Mangla, Tarun and Shaowang, Ted and Zhao, Jinjin and Paparrizos, John and Krishnan, Sanjay and Feamster, Nick},
  journal={Proceedings of the ACM on Interactive, Mobile, Wearable and Ubiquitous Technologies},
  volume={7},
  number={1},
  pages={1--26},
  year={2023},
  publisher={ACM New York, NY, USA}
}

@INPROCEEDINGS{ustc,
  author={Wei Wang and Ming Zhu and Xuewen Zeng and Xiaozhou Ye and Yiqiang Sheng},
  booktitle={2017 International Conference on Information Networking (ICOIN)}, 
  title={Malware traffic classification using convolutional neural network for representation learning}, 
  year={2017},
  volume={},
  number={},
  pages={712-717},
  doi={10.1109/ICOIN.2017.7899588}}

@INPROCEEDINGS{cic,
  author={MontazeriShatoori, Mohammadreza and Davidson, Logan and Kaur, Gurdip and Habibi Lashkari, Arash},
  booktitle={2020 IEEE Intl Conf on Dependable, Autonomic and Secure Computing, Intl Conf on Pervasive Intelligence and Computing, Intl Conf on Cloud and Big Data Computing, Intl Conf on Cyber Science and Technology Congress (DASC/PiCom/CBDCom/CyberSciTech)}, 
  title={Detection of DoH Tunnels using Time-series Classification of Encrypted Traffic}, 
  year={2020},
  volume={},
  number={},
  pages={63-70},
  doi={10.1109/DASC-PICom-CBDCom-CyberSciTech49142.2020.00026}}

@inproceedings{ganju2018property,
  title={Property inference attacks on fully connected neural networks using permutation invariant representations},
  author={Ganju, Karan and Wang, Qi and Yang, Wei and Gunter, Carl A and Borisov, Nikita},
  booktitle={Proceedings of the 2018 ACM SIGSAC conference on computer and communications security},
  pages={619--633},
  year={2018}
}

@inproceedings{fredrikson2015model,
  title={Model inversion attacks that exploit confidence information and basic countermeasures},
  author={Fredrikson, Matt and Jha, Somesh and Ristenpart, Thomas},
  booktitle={Proceedings of the 22nd ACM SIGSAC conference on computer and communications security},
  pages={1322--1333},
  year={2015}
}

@article{hayase2024data,
  title={Data mixture inference attack: BPE tokenizers reveal training data compositions},
  author={Hayase, Jonathan and Liu, Alisa and Choi, Yejin and Oh, Sewoong and Smith, Noah A},
  journal={Advances in Neural Information Processing Systems},
  volume={37},
  pages={8956--8983},
  year={2024}
}

@inproceedings{wang2024property,
  title={Property existence inference against generative models},
  author={Wang, Lijin and Wang, Jingjing and Wan, Jie and Long, Lin and Yang, Ziqi and Qin, Zhan},
  booktitle={33rd USENIX Security Symposium (USENIX Security 24)},
  pages={2423--2440},
  year={2024}
}

@article{powar2023sok,
  title={SoK: Managing risks of linkage attacks on data privacy},
  author={Powar, Jovan and Beresford, Alastair R},
  journal={Proceedings on Privacy Enhancing Technologies},
  year={2023}
}

@article{giomi2022unified,
  title={A unified framework for quantifying privacy risk in synthetic data},
  author={Giomi, Matteo and Boenisch, Franziska and Wehmeyer, Christoph and Tasn{\'a}di, Borb{\'a}la},
  journal={arXiv preprint arXiv:2211.10459},
  year={2022}
}

@article{yao2025trafficdiary,
  title={TrafficDiary: User Attribute Inference Based on Smart Home Traffic Traces},
  author={Yao, Yunhao and Hou, Jiahui and Yuan, Mu and Zhang, Haiyue and Xu, Zhengyuan and Li, Xiang-Yang},
  journal={ACM Transactions on Internet Technology},
  year={2025},
  publisher={ACM New York, NY}
}

@inproceedings{demographics,
author = {Li, Huaxin and Xu, Zheyu and Zhu, Haojin and Ma, Di and Li, Shuai and Xing, Kai},
title = {Demographics inference through Wi-Fi network traffic analysis},
year = {2016},
publisher = {IEEE Press},
url = {https://doi.org/10.1109/INFOCOM.2016.7524528},
doi = {10.1109/INFOCOM.2016.7524528},
booktitle = {IEEE INFOCOM 2016 - The 35th Annual IEEE International Conference on Computer Communications},
pages = {1–9},
numpages = {9},
location = {San Francisco, CA, USA}
}

@inproceedings{chen2014fingerprinting,
  title={OS fingerprinting and tethering detection in mobile networks},
  author={Chen, Yi-Chao and Liao, Yong and Baldi, Mario and Lee, Sung-Ju and Qiu, Lili},
  booktitle={Proceedings of the 2014 Conference on Internet Measurement Conference},
  pages={173--180},
  year={2014}
}

@inproceedings{bai2022passive,
  title={Passive OS fingerprinting on commodity switches},
  author={Bai, Sherry and Kim, Hyojoon and Rexford, Jennifer},
  booktitle={2022 IEEE 8th International Conference on Network Softwarization (NetSoft)},
  pages={264--268},
  year={2022},
  organization={IEEE}
}

@article{oh2017p,
  title={p-FP: Extraction, classification, and prediction of website fingerprints with deep learning},
  author={Oh, Se Eun and Sunkam, Saikrishna and Hopper, Nicholas},
  journal={arXiv preprint arXiv:1711.03656},
  year={2017}
}

@inproceedings{mavroudis2023adaptive,
  title={Adaptive webpage fingerprinting from tls traces},
  author={Mavroudis, Vasilios and Hayes, Jamie},
  booktitle={2023 53rd Annual IEEE/IFIP International Conference on Dependable Systems and Networks (DSN)},
  pages={445--458},
  year={2023},
  organization={IEEE}
}

@ARTICLE{10533453,
  author={Li, Ruoyu and Li, Qing and Lin, Tao and Zou, Qingsong and Zhao, Dan and Huang, Yucheng and Tyson, Gareth and Xie, Guorui and Jiang, Yong},
  journal={IEEE/ACM Transactions on Networking}, 
  title={DeviceRadar: Online IoT Device Fingerprinting in ISPs Using Programmable Switches}, 
  year={2024},
  volume={32},
  number={5},
  pages={3854-3869},
  doi={10.1109/TNET.2024.3398778}}

@article{maali2025evaluating,
  title={Evaluating Machine Learning-Based IoT Device Identification Models for Security Applications},
  author={Maali, Eman and Alrawi, Omar and McCann, Julie},
  journal={arXiv preprint},
  year={2025}
}

@inproceedings{coull2007playing,
  title={Playing Devil's Advocate: Inferring Sensitive Information from Anonymized Network Traces.},
  author={Coull, Scott E and Wright, Charles V and Monrose, Fabian and Collins, Michael P and Reiter, Michael K and others},
  booktitle={Ndss},
  volume={7},
  pages={35--47},
  year={2007}
}

@inproceedings{yao2003topology,
  title={Topology inference in the presence of anonymous routers},
  author={Yao, Bin and Viswanathan, Ramesh and Chang, Fangzhe and Waddington, Daniel},
  booktitle={IEEE INFOCOM 2003. Twenty-second Annual Joint Conference of the IEEE Computer and Communications Societies (IEEE Cat. No. 03CH37428)},
  volume={1},
  pages={353--363},
  year={2003},
  organization={IEEE}
}

@inproceedings{hardness,
author = {Acharya, H. B. and Gouda, M. G.},
title = {On the hardness of topology inference},
year = {2011},
isbn = {364217678X},
publisher = {Springer-Verlag},
address = {Berlin, Heidelberg},
booktitle = {Proceedings of the 12th International Conference on Distributed Computing and Networking},
pages = {251–262},
numpages = {12},
location = {Bangalore, India},
series = {ICDCN'11}
}

@inproceedings{10.1145/3548785.3548793,
author = {Endres, Markus and Mannarapotta Venugopal, Asha and Tran, Tung Son},
title = {Synthetic Data Generation: A Comparative Study},
year = {2022},
isbn = {9781450397094},
publisher = {Association for Computing Machinery},
address = {New York, NY, USA},
url = {https://doi.org/10.1145/3548785.3548793},
doi = {10.1145/3548785.3548793},
booktitle = {Proceedings of the 26th International Database Engineered Applications Symposium},
pages = {94–102},
numpages = {9},
location = {Budapest, Hungary},
series = {IDEAS '22}
}

@article{van2023membership,
  title={Membership inference attacks against synthetic data through overfitting detection},
  author={Van Breugel, Boris and Sun, Hao and Qian, Zhaozhi and van der Schaar, Mihaela},
  journal={arXiv preprint arXiv:2302.12580},
  year={2023}
}

@inproceedings{ammar2002prefix,
  title={Prefix-Preserving IP Address Anonymization: Measurement-based Security Evaluation and a New Cryptography-based Scheme},
  author={Ammar, JunXu JinliangFan MostafaH and Moon, SueB},
  booktitle={Proc. IEEE Int'l Conf. Network Protocols},
  year={2002}
}

@inproceedings{10.1145/1519144.1519147,
author = {Foukarakis, Michael and Antoniades, Demetres and Polychronakis, Michalis},
title = {Deep packet anonymization},
year = {2009},
isbn = {9781605584720},
publisher = {Association for Computing Machinery},
address = {New York, NY, USA},
url = {https://doi.org/10.1145/1519144.1519147},
doi = {10.1145/1519144.1519147},
booktitle = {Proceedings of the Second European Workshop on System Security},
pages = {16–21},
numpages = {6},
location = {Nuremburg, Germany},
series = {EUROSEC '09}
}

@inproceedings{slagell2006flaim,
  title={FLAIM: A Multi-level Anonymization Framework for Computer and Network Logs.},
  author={Slagell, Adam J and Lakkaraju, Kiran and Luo, Katherine},
  booktitle={LISA},
  volume={6},
  pages={3--8},
  year={2006}
}

@article{aleroud2021anonymization,
  title={Anonymization of network traces data through condensation-based differential privacy},
  author={Aleroud, Ahmed and Yang, Fan and Pallaprolu, Sai Chaithanya and Chen, Zhiyuan and Karabatis, George},
  journal={Digital Threats: Research and Practice (DTRAP)},
  volume={2},
  number={4},
  pages={1--23},
  year={2021},
  publisher={ACM New York, NY}
}

@article{osti_1836674,
title = {Anonymization of Network Traces Data through Condensation-based Differential Privacy},
author = {Aleroud, Ahmed and Yang, Fan and Pallaprolu, Sai Chaithanya and Chen, Zhiyuan and Karabatis, George},
doi = {10.1145/3425401},
journal = {Digital Threats: Research and Practice},
number = 4,
volume = 2,
place = {United States},
year = {2021},
month = {12}
}

@article{mcsherry2010differentially,
  title={Differentially-private network trace analysis},
  author={McSherry, Frank and Mahajan, Ratul},
  journal={ACM SIGCOMM Computer Communication Review},
  volume={40},
  number={4},
  pages={123--134},
  year={2010},
  publisher={ACM New York, NY, USA}
}

@inproceedings{chen2018taranet,
  title={TARANET: Traffic-analysis resistant anonymity at the network layer},
  author={Chen, Chen and Asoni, Daniele E and Perrig, Adrian and Barrera, David and Danezis, George and Troncoso, Carmela},
  booktitle={2018 IEEE European Symposium on Security and Privacy (EuroS\&P)},
  pages={137--152},
  year={2018},
  organization={IEEE}
}

@inproceedings{cai2014cs,
  title={Cs-buflo: A congestion sensitive website fingerprinting defense},
  author={Cai, Xiang and Nithyanand, Rishab and Johnson, Rob},
  booktitle={Proceedings of the 13th Workshop on Privacy in the Electronic Society},
  pages={121--130},
  year={2014}
}

@inproceedings{wang2017walkie,
  title={$\{$Walkie-Talkie$\}$: An efficient defense against passive website fingerprinting attacks},
  author={Wang, Tao and Goldberg, Ian},
  booktitle={26th USENIX Security Symposium (USENIX Security 17)},
  pages={1375--1390},
  year={2017}
}

@article{sarwar2024mycroft,
  title={MYCROFT: Towards Effective and Efficient External Data Augmentation},
  author={Sarwar, Zain and Tran, Van and Bhagoji, Arjun Nitin and Feamster, Nick and Zhao, Ben Y and Chakraborty, Supriyo},
  journal={arXiv preprint arXiv:2410.08432},
  year={2024}
}

@inproceedings{zhu2025fedmia,
  title={FedMIA: An Effective Membership Inference Attack Exploiting" All for One" Principle in Federated Learning},
  author={Zhu, Gongxi and Li, Donghao and Gu, Hanlin and Yao, Yuan and Fan, Lixin and Han, Yuxing},
  booktitle={Proceedings of the Computer Vision and Pattern Recognition Conference},
  pages={20643--20653},
  year={2025}
}

@article{kowalczuk2025privacy,
  title={Privacy attacks on image autoregressive models},
  author={Kowalczuk, Antoni and Dubi{\'n}ski, Jan and Boenisch, Franziska and Dziedzic, Adam},
  journal={arXiv preprint arXiv:2502.02514},
  year={2025}
}

@inproceedings{carlini2022membership,
  title={Membership inference attacks from first principles},
  author={Carlini, Nicholas and Chien, Steve and Nasr, Milad and Song, Shuang and Terzis, Andreas and Tramer, Florian},
  booktitle={2022 IEEE symposium on security and privacy (SP)},
  pages={1897--1914},
  year={2022},
  organization={IEEE}
}

@INPROCEEDINGS{710701,
  author={Rubner, Y. and Tomasi, C. and Guibas, L.J.},
  booktitle={Sixth International Conference on Computer Vision (IEEE Cat. No.98CH36271)}, 
  title={A metric for distributions with applications to image databases}, 
  year={1998},
  volume={},
  number={},
  pages={59-66},
  doi={10.1109/ICCV.1998.710701}}

@ARTICLE{1268075,
  author={Spring, N. and Mahajan, R. and Wetherall, D. and Anderson, T.},
  journal={IEEE/ACM Transactions on Networking}, 
  title={Measuring ISP topologies with Rocketfuel}, 
  year={2004},
  volume={12},
  number={1},
  pages={2-16},
  doi={10.1109/TNET.2003.822655}}

@inproceedings{zhang2000detecting,
  title={Detecting stepping stones.},
  author={Zhang, Yin and Paxson, Vern},
  booktitle={USENIX Security Symposium},
  volume={171},
  pages={184},
  year={2000}
}

@article{soltani2024security,
  title={Security of topology discovery service in sdn: Vulnerabilities and countermeasures},
  author={Soltani, Sanaz and Amanlou, Ali and Shojafar, Mohammad and Tafazolli, Rahim},
  journal={IEEE Open Journal of the Communications Society},
  volume={5},
  pages={3410--3450},
  year={2024},
  publisher={IEEE}
}

@misc{vnat_dataset,
  title = {{VPN-nonVPN} Network Application Traffic Dataset ({VNAT})},
  author = {{MIT Lincoln Laboratory}},
  year = {2017},
  howpublished = {\url{https://www.ll.mit.edu/r-d/datasets/vpnnonvpn-network-application-traffic-dataset-vnat}},
  note = {Accessed: 2025-11-18}
}

@inproceedings{nguyen2019diot,
  title={D{\"I}oT: A federated self-learning anomaly detection system for IoT},
  author={Nguyen, Thien Duc and Marchal, Samuel and Miettinen, Markus and Fereidooni, Hossein and Asokan, Nadarajah and Sadeghi, Ahmad-Reza},
  booktitle={2019 IEEE 39th International conference on distributed computing systems (ICDCS)},
  pages={756--767},
  year={2019},
  organization={IEEE}
}

@article{meidan2017detection,
  title={Detection of unauthorized IoT devices using machine learning techniques},
  author={Meidan, Yair and Bohadana, Michael and Shabtai, Asaf and Ochoa, Martin and Tippenhauer, Nils Ole and Guarnizo, Juan Davis and Elovici, Yuval},
  journal={arXiv preprint arXiv:1709.04647},
  year={2017}
}

@article{aziz2023content,
  title={Content-aware network traffic prediction framework for quality of service-aware dynamic network resource management},
  author={Aziz, Waqar Ali and Ioannou, Iacovos I and Lestas, Marios and Qureshi, Hassaan Khaliq and Iqbal, Adnan and Vassiliou, Vasos},
  journal={IEEE Access},
  volume={11},
  pages={99716--99733},
  year={2023},
  publisher={IEEE}
}

@article{salem2018ml,
  title={Ml-leaks: Model and data independent membership inference attacks and defenses on machine learning models},
  author={Salem, Ahmed and Zhang, Yang and Humbert, Mathias and Berrang, Pascal and Fritz, Mario and Backes, Michael},
  journal={arXiv preprint arXiv:1806.01246},
  year={2018}
}

\appendix
\section{Ethics}
This work does not raise any ethical issues.

\section{Content generated by AI}
Various tools, including Copilot, Claude, ChatGPT, and Gemini, supported the preparation of this manuscript through editing, grammar checks, and retrieval of background information. All generated text was reviewed and revised by the authors, and the core ideas, analyses, and conclusions remain entirely the authors’ own. The same tools were also used during code development and early-stage manuscript reviews.

\section{Description of four generative models used for evaluation}\label{app:models}
\textbf{NetShare: }NetShare is a GAN-based approach \cite{yin2022practical} that generates synthetic network traffic by training a generator and discriminator in adversarial fashion until synthetic samples become indistinguishable from real data. NetShare groups packets into flows based on 5-tuples and applies time-series GANs to model temporal dependencies within flows.
NetShare generates flow-level statistics and select packet header features including IP addresses, ports, protocols, TTL, packet sizes, and inter-arrival times. It encodes categorical fields (IP addresses, ports, protocols) using bitwise encoding and IP2Vec embeddings, and applies log transformation to continuous fields (packet sizes, byte counts) to reduce value ranges. The method enhances scalability through fine-tuning and improves privacy by incorporating differentially private training with a mix of public and private data.
However, NetShare has limitations: it generates only specific header fields rather than full packet-level traces with complete payloads, which are essential for applications requiring full packet details such as debugging or network provisioning.

\textbf{NetDiffusion: }Diffusion models generate data by learning to reverse a gradual noising process, transforming random noise into structured outputs through iterative denoising.
NetDiffusion \cite{jiang2024netdiffusion} applies this approach using image-based representations. The method converts packet headers into 2D binary matrices where each cell represents a single bit, then renders these matrices as images using three-value color encoding (0, 1, and missing/undefined). During training, each image is paired with a text prompt describing the traffic type (e.g., "Netflix streaming traffic") to fine-tune a pre-trained Stable Diffusion model. At generation time, the model conditions on both the text prompt and an edge map extracted via ControlNet, which provides structural guidance to maintain traffic patterns.
However, image-based representations have fundamental limitations for network traffic generation. Encoding packets as pixels obscures critical protocol semantics and inter-packet dependencies that govern valid network behavior, causing generated traffic to frequently violate protocol specifications and require post-processing corrections.

\textbf{TrafficLLM} 
Transformer-based models leverage self-attention mechanisms to capture complex relationships between tokens, enabling realistic sequence generation. For network traffic, these models tokenize packet headers into discrete units (e.g., hexadecimal values) and process them using transformer architectures.
TrafficLLM \cite{cui2025trafficllm} is a representative transformer-based approach that fine-tunes GPT-2 on tokenized traffic through a two-stage pipeline. The first stage enhances instruction-following capabilities, while the second stage improves traffic pattern representation learning. The method employs Extensible Adaptation with Parameter-Efficient Fine-Tuning (EA-PEFT) to enhance generalization across diverse instructions and datasets. 

\textbf{NetSSM}
 State-space models (SSMs) represent dynamic systems through hidden internal states that evolve over time.
NetSSM \cite{chu2025netssm} applies this framework to network traffic generation through a structured tokenization and pretraining pipeline. The approach tokenizes packet headers at the byte level, mapping each byte to the range [0, 255], and augments these sequences with control tokens: label tokens specify traffic type (e.g., $<|$Netflix$|>$), while boundary tokens ($<|$pkt$|>$) delimit individual packets. This scheme enables the model to learn both fine-grained byte patterns and higher-level packet structure. The tokenized sequences are used to pretrain an SSM architecture, where the model learns to predict subsequent tokens given preceding tokens. Once trained, generation proceeds autoregressively: conditioned on a traffic type label and an initial prompt (typically the first packet from a capture), NetSSM produces new traffic sequences token-by-token. The model can generate both single-flow traffic, where all packets share the same 5-tuple, and multi-flow scenarios containing packets from multiple distinct flows.

\newpage
\section{Additional results}

\begin{figure}[h!]
    \centering
    \includegraphics[width=0.6\linewidth]{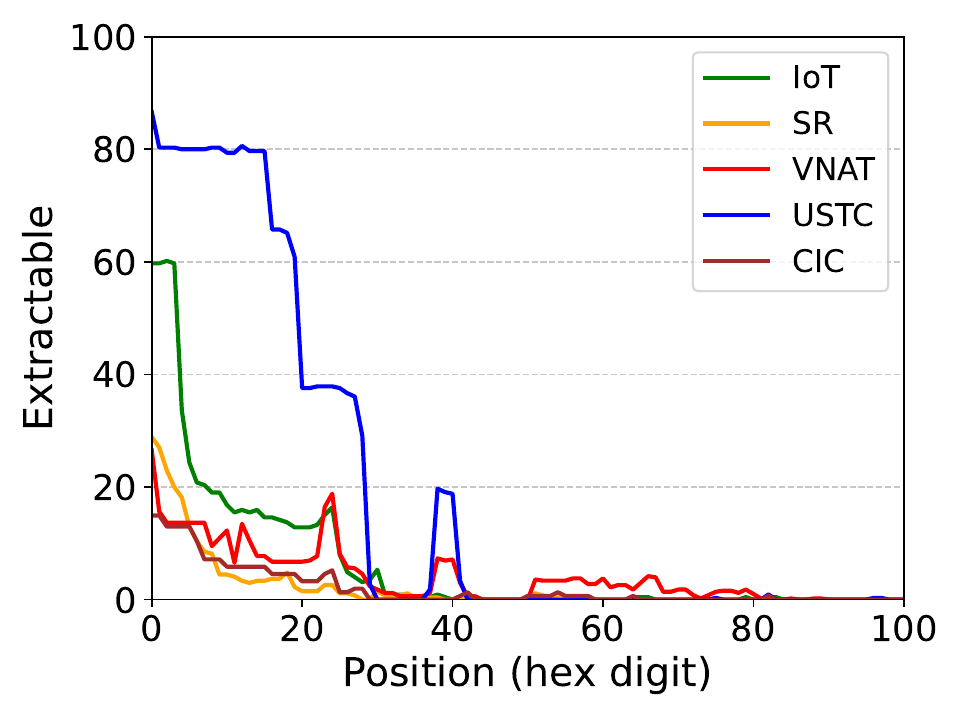}
    \vspace{-0.5em}
    \caption{Extractable rate at different token positions for TrafficLLM. Lower token positions have higher extractable rates than higher token positions.}
    \label{fig:extractable-position-TrafficLLM}
    \vspace{-0.5em}
\end{figure}

\begin{figure}[h!]
    \centering
    \begin{subfigure}[b]{0.74\linewidth}
        \centering
        \includegraphics[width=\linewidth]{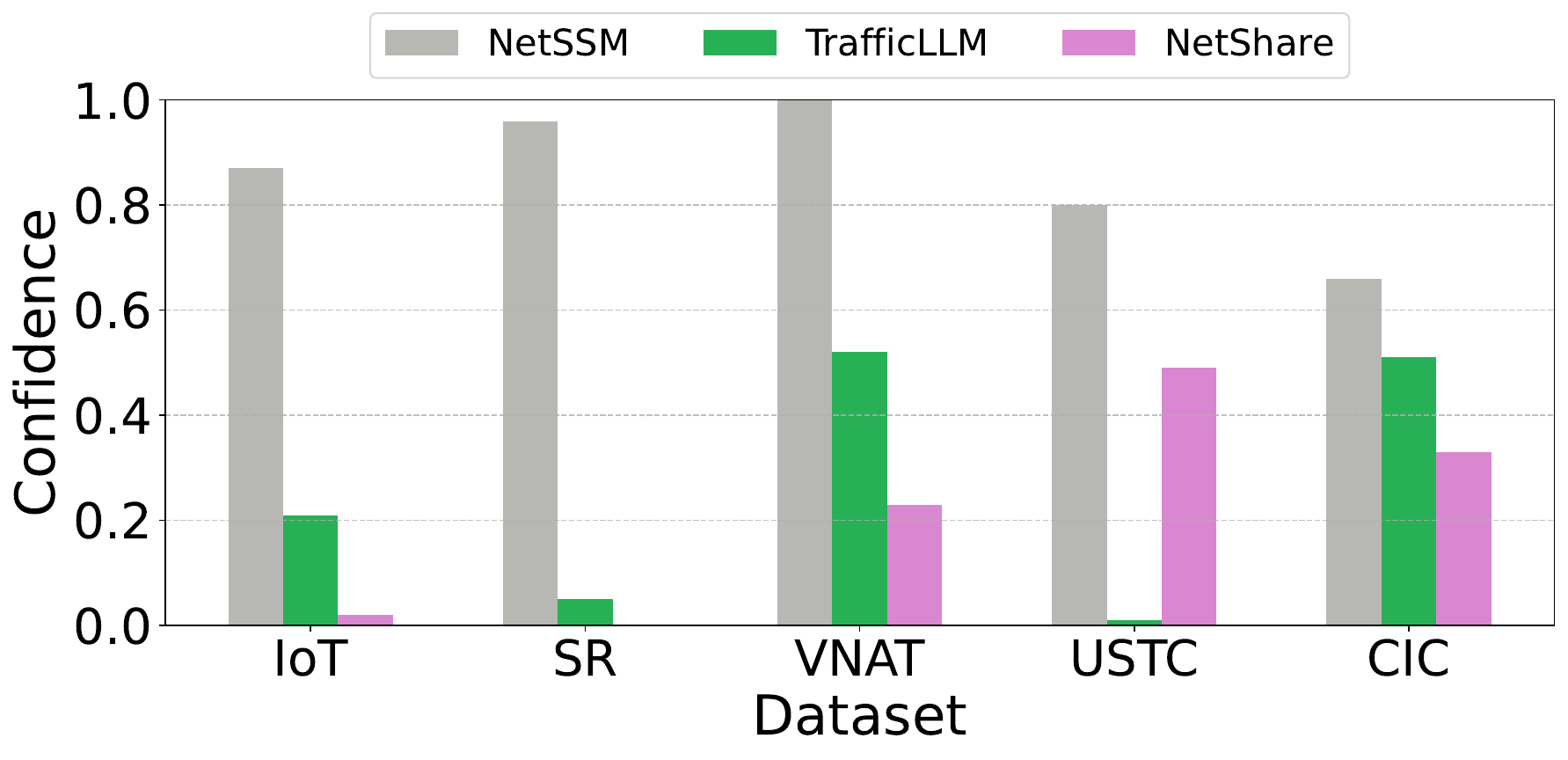}
        \caption{Confidence}
    \end{subfigure}
    \hfill
    \begin{subfigure}[b]{0.74\linewidth}
        \centering
        \includegraphics[width=\linewidth]{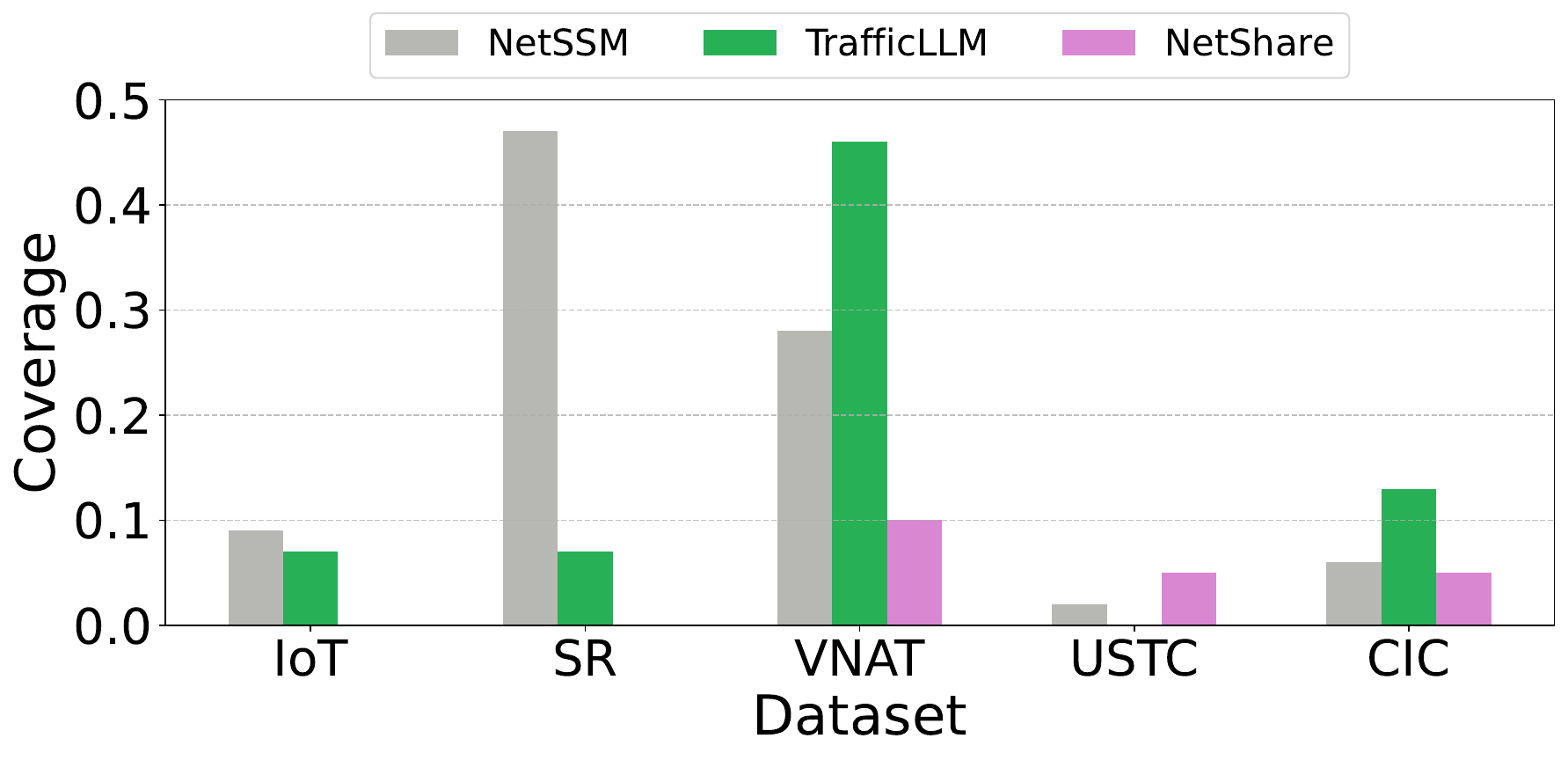}
        \caption{Coverage}
    \end{subfigure}
     \vspace{-0.5em}
    \caption{Network identifier leakage performance for destination IP addresses. We observe high confidence scores across all models, particularly in NetSSM.}
    \label{fig:leakage-destinationIP}
\end{figure}

\end{document}